\documentclass{article}
\usepackage{nips15submit_e,times}
\usepackage[small,bf]{caption}
\usepackage{hyperref}
\usepackage{url}

\usepackage{amsmath}
\usepackage{amsfonts}
\usepackage{amsthm}
\usepackage{amssymb}
\usepackage{sansmath}
\DeclareMathOperator*{\argmax}{arg\,max}

\usepackage{graphicx}
\usepackage{subfig}
\usepackage{wrapfig}

\usepackage{algorithm}
\usepackage{algorithmic}

\usepackage{booktabs}
\usepackage{relsize}
\usepackage{lipsum}

\usepackage{inconsolata}
\usepackage[T1]{fontenc}
\usepackage{color}
\usepackage{verbatim}

\def\ifmonospace{\ifdim\fontdimen3\font=0pt}

\def\C++{%
\ifmonospace%
    C++%
\else%
    C\kern-.1667em\raise.30ex\hbox{\smaller{++}}%
\fi%
\spacefactor1000 }

\title{Deep Speech 2: End-to-End Speech Recognition in English and Mandarin}
\author{Baidu Research -- Silicon Valley AI Lab\thanks{Authorship order is alphabetical.} \\ 
Dario Amodei, Rishita Anubhai, Eric Battenberg,  Carl Case, Jared Casper, Bryan Catanzaro, \\ Jingdong Chen, Mike Chrzanowski, Adam Coates, Greg Diamos, Erich Elsen, Jesse	Engel, \\ Linxi Fan, Christopher Fougner, Tony Han, Awni Hannun, Billy Jun, Patrick LeGresley, \\ Libby Lin, Sharan Narang, Andrew Ng, Sherjil Ozair, Ryan Prenger, Jonathan Raiman, \\ Sanjeev Satheesh, David Seetapun, Shubho Sengupta, Yi Wang, Zhiqian Wang, Chong Wang, \\ Bo Xiao, Dani Yogatama, Jun Zhan, Zhenyao Zhu}

\nipsfinalcopy

\begin{document} 
\maketitle
\vskip -0.2in
\begin{abstract} 

We show that an end-to-end deep learning approach can be used to recognize either English or Mandarin Chinese speech---two vastly different languages. Because it replaces entire pipelines of hand-engineered components with neural networks, end-to-end learning allows us to handle a diverse variety of speech including noisy environments, accents and different languages. Key to our approach is our application of HPC techniques, resulting in a 7x speedup over our previous system~\cite{hannun2014deepspeech}. Because of this efficiency, experiments that previously took weeks now run in days. This enables us to iterate more quickly to identify superior architectures and algorithms. As a result, in several cases, our system is competitive with the transcription of human workers when benchmarked on standard datasets. Finally, using a technique called Batch Dispatch with GPUs in the data center, we show that our system can be inexpensively deployed in an online setting, delivering low latency when serving users at scale.

\end{abstract}

\section{Introduction}
\label{section:intro}

Decades worth of hand-engineered domain knowledge has gone into current state-of-the-art automatic speech recognition (ASR) pipelines.  A simple but powerful alternative solution is to train such ASR models end-to-end, using deep learning to replace most modules with a single model~\cite{hannun2014deepspeech}. We present the second generation of our speech system that exemplifies the major advantages of end-to-end learning. The {Deep Speech 2} ASR pipeline approaches or exceeds the accuracy of Amazon Mechanical Turk human workers on several benchmarks, works in multiple languages with little modification, and is deployable in a production setting. It thus represents a significant step towards a single ASR system that addresses the entire range of speech recognition contexts handled by humans. Since our system is built on end-to-end deep learning, we can employ a spectrum of deep learning techniques: capturing large training sets, training larger models with high performance computing, and methodically exploring the space of neural network architectures. We show that through these techniques we are able to reduce error rates of our previous end-to-end system~\cite{hannun2014deepspeech} in English by up to 43\%, and can also recognize Mandarin speech with high accuracy.

One of the challenges of speech recognition is the wide range of variability in speech and acoustics. As a result, modern ASR pipelines are made up of numerous components including complex feature extraction, acoustic models, language and pronunciation models, speaker adaptation, etc. Building and tuning these individual components makes developing a new speech recognizer very hard, especially for a new language.  Indeed, many parts do not generalize well across environments or languages and it is often necessary to support multiple application-specific systems in order to provide acceptable accuracy. This state of affairs is different from human speech recognition: people have the innate ability to learn any language during childhood, using general skills to learn language. After learning to read and write, most humans can transcribe speech with robustness to variation in environment, speaker accent and noise, without additional training for the transcription task.  To meet the expectations of speech recognition users, we believe that a single engine must learn to be similarly competent; able to handle most applications with only minor modifications and able to learn new languages from scratch without dramatic changes.
Our end-to-end system puts this goal within reach, allowing us to approach or exceed the performance of human workers on several tests in two very different languages: Mandarin and English.

Since {Deep Speech 2} (DS2) is an end-to-end deep learning system, we can achieve performance gains by focusing on three crucial components:  the model architecture, large labeled training datasets, and computational scale. This approach has also yielded great advances in other application areas such as computer vision and natural language.  This paper details our contribution to these three areas for speech recognition, including an extensive investigation of model architectures and the effect of data and model size on recognition performance.  In particular, we describe numerous experiments with neural networks trained with the Connectionist Temporal Classification (CTC) loss function~\cite{graves2006} to predict speech transcriptions from audio.  We consider networks composed of many layers of recurrent connections, convolutional filters, and nonlinearities, as well as the impact of a specific instance of Batch Normalization~\cite{ioffe2015} (BatchNorm) applied to RNNs.  We not only find networks that produce much better predictions than those in previous work~\cite{hannun2014deepspeech}, but also find instances of recurrent models that can be deployed in a production setting with no significant loss in accuracy.

Beyond the search for better model architecture, deep learning systems benefit greatly from large quantities of training data.  We detail our data capturing pipeline that has enabled us to create larger datasets than what is typically used to train speech recognition systems.  Our English speech system is trained on 11,940 hours of speech, while the Mandarin system is trained on 9,400 hours.  We use data synthesis to further augment the data during training.

Training on large quantities of data usually requires the use of larger models. Indeed, our models have many more parameters than those used in our previous system. Training a single model at these scales requires tens of exaFLOPs\footnote{1 exaFLOP = $10^{18}$ FLoating-point OPerations.} that would require 3-6 weeks to execute on a single GPU. This makes model exploration a very time consuming exercise, so we have built a highly optimized training system that uses 8 or 16 GPUs to train one model. In contrast to previous large-scale training approaches that use parameter servers and asynchronous updates~\cite{dean2012largescale, chilimbi2014adam},
we use synchronous SGD, which is easier to debug while testing new ideas, and also converges faster for the same degree of data parallelism. To make the entire system efficient, we describe optimizations for a single GPU as well as improvements to scalability for multiple GPUs. We employ optimization techniques typically found in High Performance Computing to improve scalability. These optimizations include a fast implementation of the CTC loss function on the GPU, and a custom memory allocator. We also use carefully integrated compute nodes and a custom implementation of all-reduce to accelerate inter-GPU communication. Overall the system sustains approximately 50 teraFLOP/second when training on 16 GPUs. This amounts to 3 teraFLOP/second per GPU which is about 50\% of peak theoretical performance. This scalability and efficiency cuts training times down to 3 to 5 days, allowing us to iterate more quickly on our models and datasets.

We benchmark our system on several publicly available test sets and compare the results to our previous end-to-end system~\cite{hannun2014deepspeech}.  Our goal is to eventually reach human-level performance not only on specific benchmarks, where it is possible to improve through dataset-specific tuning, but on a range of benchmarks that reflects a diverse set of scenarios. To that end, we have also measured the performance of human workers on each benchmark for comparison.  We find that our system outperforms humans in some commonly-studied benchmarks and has significantly closed the gap in much harder cases.  In addition to public benchmarks, we show the performance of our Mandarin system on internal datasets that reflect real-world product scenarios.

Deep learning systems can be challenging to deploy at scale.  Large neural networks are computationally expensive to evaluate for each user utterance, and some network architectures are more easily deployed than others. Through model exploration, we find high-accuracy, deployable network architectures, which we detail here. We also employ a batching scheme suitable for GPU hardware called Batch Dispatch that leads to an efficient, real-time implementation of our Mandarin engine on production servers.  Our implementation achieves a 98th percentile compute latency of 67 milliseconds, while the server is loaded with 10 simultaneous audio streams.

The remainder of the paper is as follows. We begin with a review of related work in deep learning, end-to-end speech recognition, and scalability in Section~\ref{section:related}. Section~\ref{section:model} describes the architectural and algorithmic improvements to the model and Section~\ref{section:optimization} explains how to efficiently compute them. We discuss the training data and steps taken to further augment the training set in Section~\ref{section:data}. An analysis of results for the DS2 system in English and Mandarin is presented in Section~\ref{section:results}. We end with a description of the steps needed to deploy DS2 to real users in Section~\ref{section:deployment}.

\section{Related Work}
\label{section:related}

This work is inspired by previous work in both deep learning and speech recognition. Feed-forward neural network acoustic models were explored more than 20 years ago~\cite{bourlard93, renals1994, ellis1999}. Recurrent neural networks and networks with convolution were also used in speech recognition around the same time~\cite{robinson1996, waibel1989}. More recently DNNs have become a fixture in the ASR pipeline with almost all state of the art speech work containing some form of deep neural network~\cite{mohamed2011, hinton2012, dahl2011a,dahl2011, jaitly2012, seide2011b}. Convolutional networks have also been found beneficial for acoustic models~\cite{abdelhamid2012, sainath2013cnn}. Recurrent neural networks, typically LSTMs, are just beginning to be deployed in state-of-the art recognizers~\cite{graves2013drnn, sak2014, sak2014b} and work well together with convolutional layers for the feature extraction~\cite{sainath2015}.  Models with both bidirectional~\cite{graves2013drnn} and unidirectional recurrence have been explored as well.

End-to-end speech recognition is an active area of research, showing compelling results when used to re-score the outputs of a DNN-HMM~\cite{graves2014} and standalone~\cite{hannun2014deepspeech}. Two methods are currently used to map variable length audio sequences directly to variable length transcriptions. The RNN encoder-decoder paradigm uses an encoder RNN to map the input to a fixed length vector and a decoder network to expand the fixed length vector into a sequence of output predictions~\cite{cho2014, sutskever2014seq}. Adding an attentional mechanism to the decoder greatly improves performance of the system, particularly with long inputs or outputs~\cite{bahdanau2015}. In speech, the RNN encoder-decoder with attention performs well both in predicting phonemes~\cite{chorowski2015firstresults} or graphemes~\cite{bahdanau2015b, chan2015}. 

The other commonly used technique for mapping variable length audio input to variable length output is the CTC loss function~\cite{graves2006} coupled with an RNN to model temporal information. The CTC-RNN model performs well in end-to-end speech recognition with grapheme outputs~\cite{graves2014, hannun2014firstpass, hannun2014deepspeech, maas2015}. The CTC-RNN model has also been shown to work well in predicting phonemes~\cite{miao2015, sak2015}, though a lexicon is still needed in this case.
Furthermore it has been necessary to pre-train the CTC-RNN network with a DNN cross-entropy network that is fed frame-wise alignments from a GMM-HMM system~\cite{sak2015}. In contrast, we train the CTC-RNN networks from scratch without the need of frame-wise alignments for pre-training.

Exploiting scale in deep learning has been central to the success of the field thus far~\cite{krizhevsky2012imagenet, le2012faces}. Training on a single GPU resulted in substantial performance gains~\cite{raina2009large}, which were subsequently scaled linearly to two~\cite{krizhevsky2012imagenet} or more GPUs~\cite{coates2013cotshpc}. We take advantage of work in increasing individual GPU efficiency for low-level deep learning primitives~\cite{chetlur14}. We build on the past work in using model-parallelism~\cite{coates2013cotshpc}, data-parallelism~\cite{dean2012largescale} or a combination of the two~\cite{szegedy2014googlenet, hannun2014deepspeech} to create a fast and highly scalable system for training deep RNNs in speech recognition.

Data has also been central to the success of end-to-end speech recognition, with over 7000 hours of labeled speech used in Deep Speech 1 (DS1)~\cite{hannun2014deepspeech}. Data augmentation has been highly effective in improving the performance of deep learning in computer vision~\cite{lecun2004learningmethods, sapp2008synth, coates2011icdar}. This has also been shown to improve speech systems~\cite{gales2009, hannun2014deepspeech}. Techniques used for data augmentation in speech range from simple noise addition~\cite{hannun2014deepspeech} to complex perturbations such as simulating changes to the vocal tract length and rate of speech of the speaker~\cite{jaitly2013, ko2015}. 

Existing speech systems can also be used to bootstrap new data collection. In one approach, the authors use one speech engine to align and filter a thousand hours of read speech~\cite{panayotov2015}. In another approach, a heavy-weight offline speech recognizer is used to generate transcriptions for tens of thousands of hours of speech~\cite{kapralove2014}. This is then passed through a filter and used to re-train the recognizer, resulting in significant performance gains. We draw inspiration from these past approaches in bootstrapping larger datasets and data augmentation to increase the effective amount of labeled data for our system.

\section{Model Architecture}
\label{section:model}

A simple multi-layer model with a single recurrent layer cannot exploit thousands of hours of labelled speech. In order to learn from datasets this large, we increase the model capacity via depth. We explore architectures with up to 11 layers including many bidirectional recurrent layers and convolutional layers. These models have nearly 8 times the amount of computation per data example as the models in Deep Speech 1 making fast optimization and computation critical. In order to optimize these models successfully, we use Batch Normalization for RNNs and a novel optimization curriculum we call SortaGrad. We also exploit long \emph{strides} between RNN inputs to reduce computation per example by a factor of 3. This is helpful for both training and evaluation, though requires some modifications in order to work well with CTC. Finally, though many of our research results make use of bidirectional recurrent layers, we find that excellent models exist using only unidirectional recurrent layers---a feature that makes such models much easier to deploy.  Taken together these features allow us to tractably optimize deep RNNs and improve performance by more than 40\% in both English and Mandarin error rates over the smaller baseline models.

\subsection{Preliminaries}

\begin{figure}[h]
    \centering
    \includegraphics[width=0.6\textwidth]{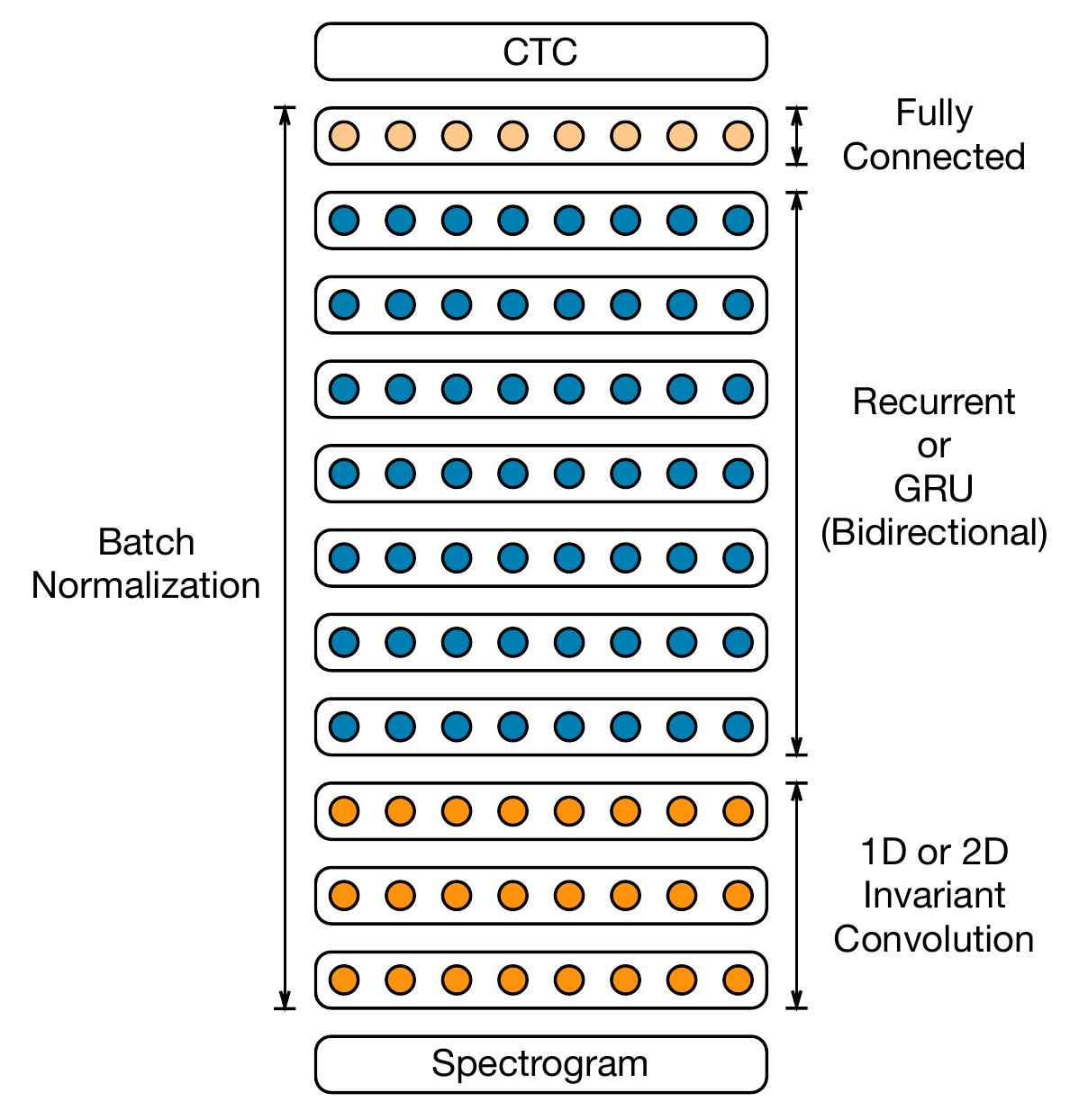}
    \caption{Architecture of the DS2 system used to train on both English and Mandarin speech. We explore variants of this architecture by varying the number of convolutional layers from 1 to 3 and the number of recurrent or GRU layers from 1 to 7.}
    \label{fig:ds2-network}
\end{figure}

Figure~\ref{fig:ds2-network} shows the architecture of the DS2 system which at its core is similar to the previous DS1 system~\cite{hannun2014deepspeech}: a recurrent neural network (RNN) trained to ingest speech spectrograms and generate text transcriptions. 

Let a single utterance $x^{(i)}$ and label $y^{(i)}$ be sampled from a training set $\mathcal{X} = \{(x^{(1)},y^{(1)}),(x^{(2)},y^{(2)}),\ldots\}$. Each utterance, $x^{(i)}$, is a time-series of length $T^{(i)}$ where every time-slice is a vector of audio features, $x_t^{(i)},  t=0,\ldots,T^{(i)}-1$. We use a spectrogram of power normalized audio clips as the features to the system, so $x^{(i)}_{t,p}$ denotes the power of the $p$'th frequency bin in the audio frame at time $t$. The goal of the RNN is to convert an input sequence $x^{(i)}$ into a final transcription $y^{(i)}$. For notational convenience, we drop the superscripts and use $x$ to denote a chosen utterance and $y$ the corresponding label.

The outputs of the network are the graphemes of each language. At each output time-step $t$, the RNN makes a prediction over characters, $p(\ell_t | x)$, where $\ell_t$ is either a character in the alphabet or the blank symbol. In English we have $\ell_t \in \{\textrm{a, b, c, }\ldots, \textrm{z}, \textit{space}, \textit{apostrophe}, \textit{blank}\}$, where we have added the \textit{apostrophe} as well as a \textit{space} symbol to denote word boundaries. For the Mandarin system the network outputs simplified Chinese characters.  We describe this in more detail in Section~\ref{subsection:chinesemodel}.

The RNN model is composed of several layers of hidden units. The architectures we experiment with consist of one or more convolutional layers, followed by one or more recurrent layers, followed by one or more fully connected layers.

The hidden representation at layer $l$ is given by $h^l$ with the convention that $h^0$ represents the input $x$. The bottom of the network is one or more convolutions over the time dimension of the input. For a context window of size $c$, the $i$-th activation at time-step $t$ of the convolutional layer is given by
\begin{equation}
h_{t,i}^l = f( w_i^l \circ h^{l-1}_{t-c:t+c} )
\end{equation}
where $\circ$ denotes the element-wise product between the $i$-th filter and the context window of the previous layers activations, and $f$ denotes a unary nonlinear function. We use the clipped rectified-linear (ReLU) function $\sigma(x) = \min\{\max\{x, 0\},20\}$ as our nonlinearity. In some layers, usually the first, we sub-sample by striding the convolution by $s$ frames. The goal is to shorten the number of time-steps for the recurrent layers above.

Following the convolutional layers are one or more bidirectional recurrent layers~\cite{schuster1997bidirectional}. The forward in time $\overrightarrow{h}^l$ and backward in time $\overleftarrow{h}^l$ recurrent layer activations are computed as
\begin{equation}
\begin{aligned}
    \overrightarrow{h}_t^l &= g( h_t^{l-1}, \overrightarrow{h}_{t-1}^l ) \\
    \overleftarrow{h}_t^l &= g( h_t^{l-1}, \overleftarrow{h}_{t+1}^l )
\end{aligned}
\end{equation}
The two sets of activations are summed to form the output activations for the layer $h^l = \overrightarrow{h}^l + \overleftarrow{h}^l$.
The function $g(\cdot)$ can be the standard recurrent operation
\begin{equation}
\label{eq:plainrnn}
    \overrightarrow{h}_t^l = f( W^l h_t^{l-1} + \overrightarrow{U}^l \overrightarrow{h}_{t-1}^l + b^l )
\end{equation}
where $W^l$ is the input-hidden weight matrix, $\overrightarrow{U}^l$ is the recurrent weight matrix and $b^l$ is a bias term. In this case the input-hidden weights are shared for both directions of the recurrence. The function $g(\cdot)$ can also represent more complex recurrence operations such as the Long Short-Term Memory (LSTM) units~\cite{hochreiter1997lstm} and the gated recurrent units (GRU)~\cite{cho2014}.

After the bidirectional recurrent layers we apply one or more fully connected layers with
\begin{equation}
    h^l_t = f( W^l h^{l-1}_t + b^l )
\end{equation}
The output layer $L$ is a softmax computing a probability distribution over characters given by
\begin{equation}
    p(\ell_t=k | x) = \frac{\exp(w^L_k \cdot h^{L-1}_t)}{\sum_j \exp(w^L_j \cdot h^{L-1}_t)}
\end{equation}

The model is trained using the CTC loss function~\cite{graves2006}. Given an input-output pair $(x, y)$ and the current parameters of the network $\theta$, we compute the loss function $\mathcal{L}(x, y; \theta)$ and its derivative with respect to the parameters of the network $\nabla_\theta \mathcal{L}(x, y; \theta)$. This derivative is then used to update the network parameters through the backpropagation through time algorithm.

In the following subsections we describe the architectural and algorithmic improvements made relative to DS1~\cite{hannun2014deepspeech}.  Unless otherwise stated these improvements are language agnostic. We report results on an English speaker held out development set which is an internal dataset containing 2048 utterances of primarily read speech. All models are trained on datasets described in Section~\ref{section:data}. We report Word Error Rate (WER) for the English system and Character Error Rate (CER) for the Mandarin system. In both cases we integrate a language model in a beam search decoding step as described in Section~\ref{subsection:languagemodel}.

\subsection{Batch Normalization for Deep RNNs}
\label{subsection:depth}

To efficiently scale our model as we scale the training set, we increase the depth of the networks by adding more hidden layers, rather than making each layer larger. Previous work has examined doing so by increasing the number of consecutive bidirectional recurrent layers~\cite{graves2013drnn}. We explore Batch Normalization (BatchNorm) as a technique to accelerate training for such networks~\cite{ioffe2015} since they often suffer from optimization issues. 

Recent research has shown that BatchNorm improves the speed of convergence of recurrent nets, without showing any improvement in generalization performance~\cite{laurent2015}. In contrast, we demonstrate that when applied to very deep networks of simple RNNs on large data sets, batch normalization substantially improves final generalization error while greatly accelerating training. 

In a typical feed-forward layer containing an affine transformation followed by a non-linearity $f(\cdot)$, we insert a BatchNorm transformation by applying $f(\mathcal{B}(Wh))$ instead of $f(Wh + b)$, where
\begin{equation}
    \mathcal{B}(x) = \gamma \frac{x - \mathrm{E}[x]}{\left(\mathrm{Var}[x] + \epsilon\right)^{1/2}} + \beta.
\end{equation}
The terms $\mathrm{E}$ and $\mathrm{Var}$ are the empirical mean and variance over a minibatch. The bias $b$ of the layer is dropped since its effect is cancelled by mean removal. The learnable parameters $\gamma$ and $\beta$ allow the layer to scale and shift each hidden unit as desired. The constant $\epsilon$ is small and positive, and is included only for numerical stability. In our convolutional layers the mean and variance are estimated over all the temporal output units for a given convolutional filter on a minibatch. The BatchNorm transformation reduces \emph{internal covariate shift} by insulating a given layer from potentially uninteresting changes in the mean and variance of the layer's input.

We consider two methods of extending BatchNorm to bidirectional RNNs~\cite{laurent2015}. A natural extension is to insert a BatchNorm transformation immediately before every non-linearity. Equation~\ref{eq:plainrnn} then becomes 
\begin{equation}
    \overrightarrow{h}^l_t = f( \mathcal{B}( W^l h^{l-1}_t + \overrightarrow{U}^l \overrightarrow{h}^l_{t-1} )).  
\end{equation}
In this case the mean and variance statistics are accumulated over a single time-step of the minibatch. The sequential dependence between time-steps prevents averaging over all time-steps. We find that this technique does not lead to improvements in optimization. We also tried accumulating an average over successive time-steps, so later time-steps are normalized over all present and previous time-steps. This also proved ineffective and greatly complicated backpropagation.

We find that \emph{sequence-wise} normalization~\cite{laurent2015} overcomes these issues. The recurrent computation is given by
\begin{equation}
    \overrightarrow{h}^l_t = f( \mathcal{B}( W^l h^{l-1}_t) + \overrightarrow{U}^l \overrightarrow{h}^l_{t-1}).  
\end{equation}

\begin{table}
\centering
\begin{tabular}{l  c  r r r  r  r r r  r}
\toprule
Architecture & Hidden Units & \multicolumn{4}{c}{Train} & \multicolumn{4}{c}{Dev}  \\
\midrule
     &  & \multicolumn{3}{c}{Baseline} & BatchNorm & \multicolumn{3}{c}{Baseline} & BatchNorm \\
\midrule
1 RNN, 5 total   & 2400 & & 10.55 & & 11.99 & & 13.55 & & 14.40 \\
3 RNN, 5 total   & 1880 & & 9.55  & & 8.29  & & 11.61 & & 10.56 \\
5 RNN, 7 total   & 1510 & & 8.59  & & 7.61  & & 10.77 & & 9.78 \\
7 RNN, 9 total   & 1280 & & 8.76  & & 7.68  & & 10.83 & & 9.52 \\
\bottomrule
\end{tabular}
\caption{Comparison of WER on a training and development set for various depths of RNN, with and without BatchNorm. The number of parameters is kept constant as the depth increases, thus the number of hidden units per layer decreases. All networks have 38 million parameters. The architecture ``M RNN, N total'' implies 1 layer of 1D convolution at the input, M consecutive bidirectional RNN layers, and the rest as fully-connected layers with N total layers in the network.}
\label{table:batch_norm}
\end{table}

\begin{figure}
\centering
\includegraphics[width=0.6\textwidth]{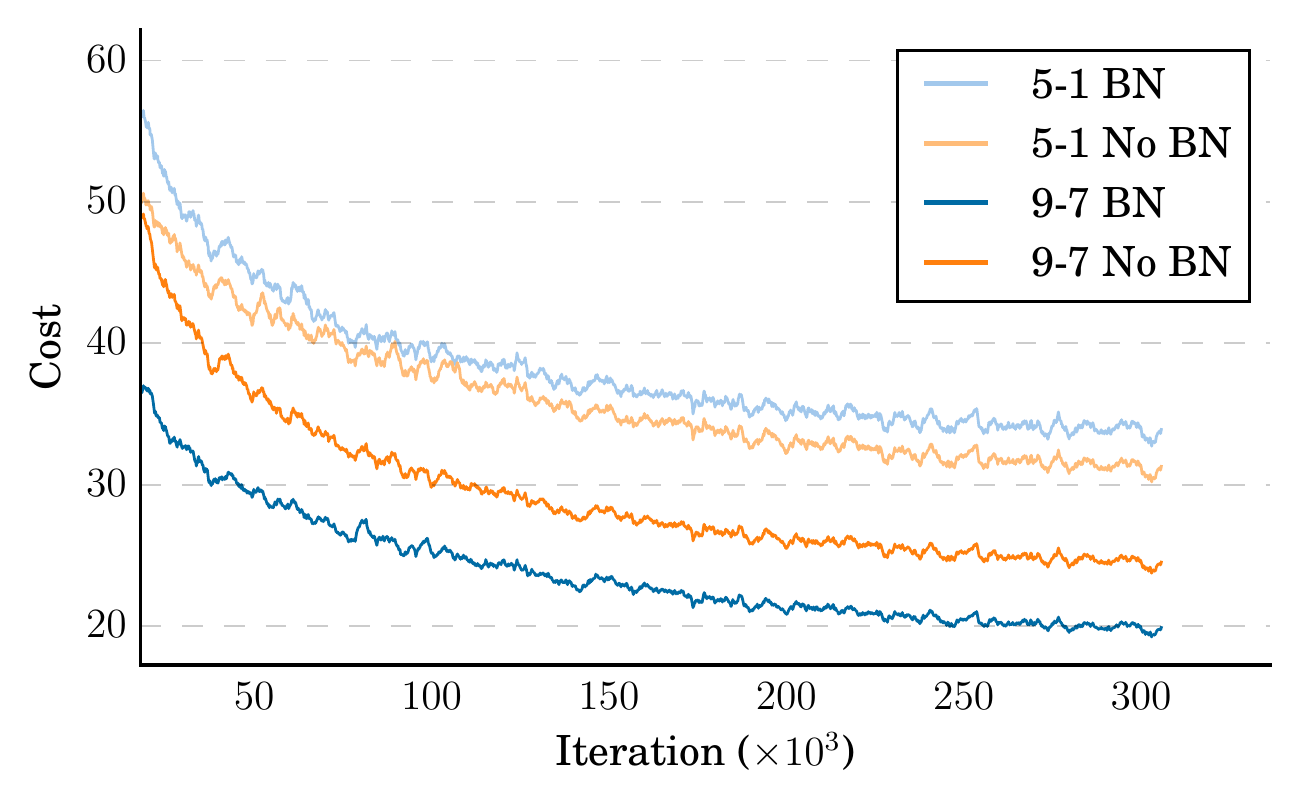}
\caption{Training curves of two models trained with and without BatchNorm. We start the plot after the first epoch of training as the curve is more difficult to interpret due to the SortaGrad curriculum method mentioned in Section~\ref{subsection:sorting}}
\label{fig:bn}
\end{figure}

For each hidden unit, we compute the mean and variance statistics over all items in the minibatch over the length of the sequence. Figure~\ref{fig:bn} shows that deep networks converge faster with sequence-wise normalization. Table~\ref{table:batch_norm} shows that the performance improvement from sequence-wise normalization increases with the depth of the network, with a 12\% performance difference for the deepest network. When comparing depth, in order to control for model size we hold constant the total number of parameters and still see strong performance gains. We would expect to see even larger improvements from depth if we held constant the number of activations per layer and added layers. We also find that BatchNorm harms generalization error for the shallowest network just as it converges slower for shallower networks. 

The BatchNorm approach works well in training, but is difficult to implement for a deployed ASR system, since it is often necessary to evaluate a single utterance in deployment rather than a batch. We find that normalizing each neuron to its mean and variance over just the sequence degrades performance. Instead, we store a running average of the mean and variance for the neuron collected during training, and use these for evaluation in deployment~\cite{ioffe2015}. Using this technique, we can evaluate a single utterance at a time with better results than evaluating with a large batch.

\subsection{SortaGrad}
\label{subsection:sorting}

Training on examples of varying length pose some algorithmic challenges. One possible solution is truncating backpropagation through time~\cite{williams1990}, so that all examples have the same sequence length during training~\cite{sainath2015}. However, this can inhibit the ability to learn longer term dependencies. Other works have found that presenting examples in order of difficulty can accelerate online learning~\cite{bengio2009curriculum, zaremba2014}. A common theme in many sequence learning problems including machine translation and speech recognition is that longer examples tend to be more challenging~\cite{cho2014}.

The CTC cost function that we use implicitly depends on the length of the utterance,
\begin{equation}
\label{eq:ctc}
    \mathcal{L}(x, y; \theta) = -\log \sum_{\ell \in \textrm{Align}(x, y)} \prod_t^{T} p_{\textrm{ctc}}(\ell_t | x; \theta).
\end{equation}
where $\textrm{Align}(x, y)$ is the set of all possible alignments of the characters of the transcription $y$ to frames of input $x$ under the CTC operator. In equation~\ref{eq:ctc}, the inner term is a product over time-steps of the sequence, which shrinks with the length of the sequence since $p_{\textrm{ctc}}(\ell_t|x;\theta)<1$. This motivates a curriculum learning strategy we title SortaGrad. SortaGrad uses the length of the utterance as a heuristic for difficulty, since long utterances have higher cost than short utterances.

\begin{table}
\centering
\begin{tabular}{l  r r r  r  r r r  r}
\toprule
& \multicolumn{4}{c}{Train} & \multicolumn{4}{c}{Dev}\\
\midrule
& \multicolumn{3}{c}{Baseline} &  BatchNorm & \multicolumn{3}{c}{Baseline} & BatchNorm\\
\midrule
Not Sorted & & 10.71 & & 8.04 & & 11.96 & & 9.78 \\
Sorted     & & 8.76 & & 7.68 & & 10.83 & & 9.52 \\
\bottomrule
\end{tabular}
\caption{Comparison of WER on a training and development set with and without SortaGrad, and with and without batch normalization.}
\label{table:sorting}
\end{table}

In the first training epoch, we iterate through the training set in increasing order of the length of the longest utterance in the minibatch. After the first epoch, training reverts back to a random order over minibatches. Table~\ref{table:sorting} shows a comparison of training cost with and without SortaGrad on the 9 layer model with 7 recurrent layers. This effect is particularly pronounced for networks without BatchNorm, since they are numerically less stable. In some sense the two techniques substitute for one another, though we still find gains when applying SortaGrad and BatchNorm together. Even with BatchNorm we find that this curriculum improves numerical stability and sensitivity to small changes in training. Numerical instability can arise from different transcendental function implementations in the CPU and the GPU, especially when computing the CTC cost. This curriculum gives comparable results for both implementations.

We suspect that these benefits occur primarily because long utterances tend to have larger gradients, yet we use a fixed learning rate independent of utterance length. Furthermore, longer utterances are more likely to cause the internal state of the RNNs to explode at an early stage in training.

\subsection{Comparison of simple RNNs and GRUs}
\label{subsection:recurrent}

The models we have shown so far are \emph{simple} RNNs that have bidirectional recurrent layers with the recurrence for both the forward in time and backward in time directions modeled by Equation~\ref{eq:plainrnn}. Current research in speech and language processing has shown that having a more complex recurrence can allow the network to remember state over more time-steps while making them more computationally expensive to train~\cite{sainath2015, chan2015, sutskever2014seq, bahdanau2015}. Two commonly used recurrent architectures are the Long Short-Term Memory (LSTM) units~\cite{hochreiter1997lstm} and the Gated Recurrent Units (GRU)~\cite{cho2014}, though many other variations exist. A recent comprehensive study of thousands of variations of LSTM and GRU architectures showed that a GRU is comparable to an LSTM with a properly initialized forget gate bias, and their best variants are competitive with each other~\cite{jozefowicz2015}. We decided to examine GRUs because experiments on smaller data sets showed the GRU and LSTM reach similar accuracy for the same number of parameters, but the GRUs were faster to train and less likely to diverge. 

The GRUs we use are computed by 
\begin{equation}
\begin{aligned}
z_t &= \sigma(W_z x_t + U_z h_{t-1} + b_z) \\
r_t &= \sigma(W_r x_t + U_r h_{t-1} + b_r) \\
\tilde{h}_t &= f(W_h x_t + r_t \circ U_h h_{t-1} + b_h) \\
h_t &= (1 - z_t) h_{t-1} + z_t \tilde{h}_t
\end{aligned}
\end{equation}

where $\sigma(\cdot)$ is the sigmoid function, $z$ and $r$ represent the \emph{update} and \emph{reset} gates respectively, and we drop the layer superscripts for simplicity. We differ slightly from the standard GRU in that we multiply the hidden state $h_{t-1}$ by $U_h$ prior to scaling by the reset gate. This allows for all operations on $h_{t-1}$ to be computed in a single matrix multiplication. The output nonlinearity $f(\cdot)$ is typically the hyperbolic tangent function $tanh$. However, we find similar performance for $tanh$ and clipped-ReLU nonlinearities and choose to use the clipped-ReLU for simplicity and uniformity with the rest of the network.

\begin{table}
\centering
\begin{tabular}{l  r  r}
\toprule
Architecture &  Simple RNN & GRU \\
\midrule
5 layers, 1 Recurrent   & 14.40 & 10.53  \\
5 layers, 3 Recurrent   & 10.56 & 8.00  \\
7 layers, 5 Recurrent   & 9.78 & 7.79  \\
9 layers, 7 Recurrent   & 9.52 & 8.19  \\
\bottomrule
\end{tabular}
\caption{Comparison of development set WER for networks with either simple RNN or GRU, for various depths. All models have batch normalization, one layer of 1D-invariant convolution, and approximately 38 million parameters.}
\label{table:rnns}
\end{table}

Both GRU and simple RNN architectures benefit from batch normalization and show strong results with deep networks. However, Table~\ref{table:rnns} shows that for a fixed number of parameters, the GRU architectures achieve better WER for all network depths. This is clear evidence of the long term dependencies inherent in the speech recognition task present both within individual words and between words. As we discuss in Section~\ref{subsection:languagemodel}, even simple RNNs are able to implicitly learn a language model due to the large amount of training data. Interestingly, the GRU networks with 5 or more recurrent layers do not significantly improve performance. We attribute this to the thinning from 1728 hidden units per layer for 1 recurrent layer to 768 hidden units per layer for 7 recurrent layers, to keep the total number of parameters constant.

The GRU networks outperform the simple RNNs in Table~\ref{table:rnns}. However, in later results (Section~\ref{section:results}) we find that as we scale up the model size, for a fixed computational budget the simple RNN networks perform slightly better. Given this, most of the remaining experiments use the simple RNN layers rather than the GRUs.

\subsection{Frequency Convolutions}
\label{subsection:2dconv}

Temporal convolution is commonly used in speech recognition to efficiently model temporal translation invariance for variable length utterances. This type of convolution was first proposed for neural networks in speech more than 25 years ago~\cite{waibel1989}. Many neural network speech models have a first layer that processes input frames with some context window~\cite{dahl2011, vesely2013}. This can be viewed as a temporal convolution with a stride of one. 

Additionally, sub-sampling is essential to make recurrent neural networks computationally tractable with high sample-rate audio. The DS1 system accomplished this through the use of a spectrogram as input and temporal convolution in the first layer with a stride parameter to reduce the number of time-steps~\cite{hannun2014deepspeech}.

Convolutions in frequency and time domains, when applied to the spectral input features prior to any other processing, can slightly improve ASR performance~\cite{abdelhamid2012, sainath2013cnn, soltau2014}. Convolution in frequency attempts to model spectral variance due to speaker variability more concisely than what is possible with large fully connected networks. Since spectral ordering of features is removed by fully-connected and recurrent layers, frequency convolutions work better as the first layers of the network.

We experiment with adding between one and three layers of convolution. These are both in the time-and-frequency domain (2D invariance) and in the time-only domain (1D invariance). In all cases we use a \emph{same} convolution, preserving the number of input features in both frequency and time. In some cases, we specify a stride across either dimension which reduces the size of the output. We do not explicitly control for the number of parameters, since convolutional layers add a small fraction of parameters to our networks. All networks shown in Table~\ref{table:2dconv} have about 35 million parameters.

We report results on two datasets---a development set of 2048 utterances (``Regular Dev'') and a much noisier dataset of 2048 utterances (``Noisy Dev'') randomly sampled from the CHiME 2015 development datasets~\cite{barker2015chime}. We find that multiple layers of 1D-invariant convolutions provides a very small benefit. The 2D-invariant convolutions improve results substantially on noisy data, while providing a small benefit on clean data. The change from one layer of 1D-invariant convolution to three layers of 2D-invariant convolution improves WER by 23.9\% on the noisy development set.

\begin{table}
\centering
\begin{tabular}{l l l l c c c}
\toprule
Architecture & Channels      & Filter dimension    & Stride        & Regular Dev & Noisy Dev \\
\midrule
1-layer 1D   & 1280          & 11                  & 2             & 9.52        & 19.36 \\
2-layer 1D   & 640, 640      & 5, 5                & 1, 2          & 9.67        & 19.21 \\
3-layer 1D   & 512, 512, 512 & 5, 5, 5             & 1, 1, 2       & 9.20        & 20.22 \\
1-layer 2D   & 32            & 41x11               & 2x2           & 8.94        & 16.22 \\
2-layer 2D   & 32, 32        & 41x11, 21x11        & 2x2, 2x1      & 9.06        & 15.71 \\
3-layer 2D   & 32, 32, 96    & 41x11, 21x11, 21x11 & 2x2, 2x1, 2x1 & 8.61        & 14.74 \\
\bottomrule
\end{tabular}
\caption{Comparison of WER for various arrangements of convolutional layers. In all cases, the convolutions are followed by 7 recurrent layers and 1 fully connected layer. For 2D-invariant convolutions the first dimension is frequency and the second dimension is time. All models have BatchNorm, SortaGrad, and 35 million parameters.}
\label{table:2dconv}
\end{table}

\subsection{Striding}

In the convolutional layers, we apply a longer stride and wider context to speed up training as fewer time-steps are required to model a given utterance. Downsampling the input sound (through FFT and convolutional striding) reduces the number of time-steps and computation required in the following layers, but at the expense of reduced performance. 

In our Mandarin models, we employ striding in the straightforward way. However, in English, striding can reduce accuracy simply because the output of our network requires at least one time-step per output character, and the number of characters in English speech per time-step is high enough to cause problems when striding\footnote{Chinese characters are more similar to English syllables than English characters. This is reflected in our training data, where there are on average 14.1 characters/s in English, while only 3.3 characters/s in Mandarin. Conversely, the Shannon entropy per character as calculated from occurrence in the training set, is less in English due to the smaller character set---4.9 bits/char compared to 12.6 bits/char in Mandarin. This implies that spoken Mandarin has a lower temporal entropy density, $\sim$41 bits/s compared to $\sim$58 bits/s, and can thus more easily be temporally compressed without losing character information.}. To overcome this, we can enrich the English alphabet with symbols representing alternate labellings like whole words, syllables or non-overlapping \emph{n}-grams. In practice, we use non-overlapping bi-graphemes or bigrams, since these are simple to construct, unlike syllables, and there are few of them compared to alternatives such as whole words. We transform unigram labels into bigram labels through a simple isomorphism.

Non-overlapping bigrams shorten the length of the output transcription and thus allow for a decrease in the length of the unrolled RNN. The sentence \emph{the cat sat} with non-overlapping bigrams is segmented as $[th, e, space, ca, t, space, sa, t ]$. Notice that for words with odd number of characters, the last character becomes an unigram and $space$ is treated as an unigram as well. This isomorphism ensures that the same words are always composed of the same bigram and unigram tokens. The output set of bigrams consists of all bigrams that occur in the training set. 

In Table~\ref{table:bigrams} we show results for both the bigram and unigram systems for various levels of striding, with or without a language model. We observe that bigrams allow for larger strides without any sacrifice in in the word error rate. This allows us to reduce the number of time-steps of the unrolled RNN benefiting both computation and memory usage.

\begin{table}
\centering
\begin{tabular}{c  r r r  r r r  r r r  r r r}
\toprule
& \multicolumn{6}{c}{Dev no LM} & \multicolumn{6}{c}{Dev LM}\\
\midrule
Stride  & \multicolumn{3}{c}{Unigrams} & \multicolumn{3}{c}{Bigrams} & \multicolumn{3}{c}{Unigrams} & \multicolumn{3}{c}{Bigrams} \\
\midrule
2 & & 14.93 & & & 14.56 & & & 9.52  & & & 9.66  & \\
3 & & 15.01 & & & 15.60 & & & 9.65  & & & 10.06 & \\
4 & & 18.86 & & & 14.84 & & & 11.92 & & & 9.93  & \\
\bottomrule
\end{tabular}
\rule{0pt}{2.6ex}
\caption{Comparison of WER with different amounts of striding for unigram and bigram outputs on a model with 1 layer of 1D-invariant convolution, 7 recurrent layers, and 1 fully connected layer. All models have BatchNorm, SortaGrad, and 35 million parameters. The models are compared on a development set with and without the use of a 5-gram language model.}
\label{table:bigrams}
\end{table}

\subsection{Row Convolution and Unidirectional Models}
\label{section:fom}

Bidirectional RNN models are challenging to deploy in an online, low-latency setting, because they are built to operate on an entire sample, and so it is not possible to perform the transcription process as the utterance streams from the user. We have found an unidirectional architecture that performs as well as our bidirectional models. This allows us to use unidirectional, forward-only RNN layers in our deployment system. 

\begin{figure}
\centering
\includegraphics[width=0.6\textwidth]{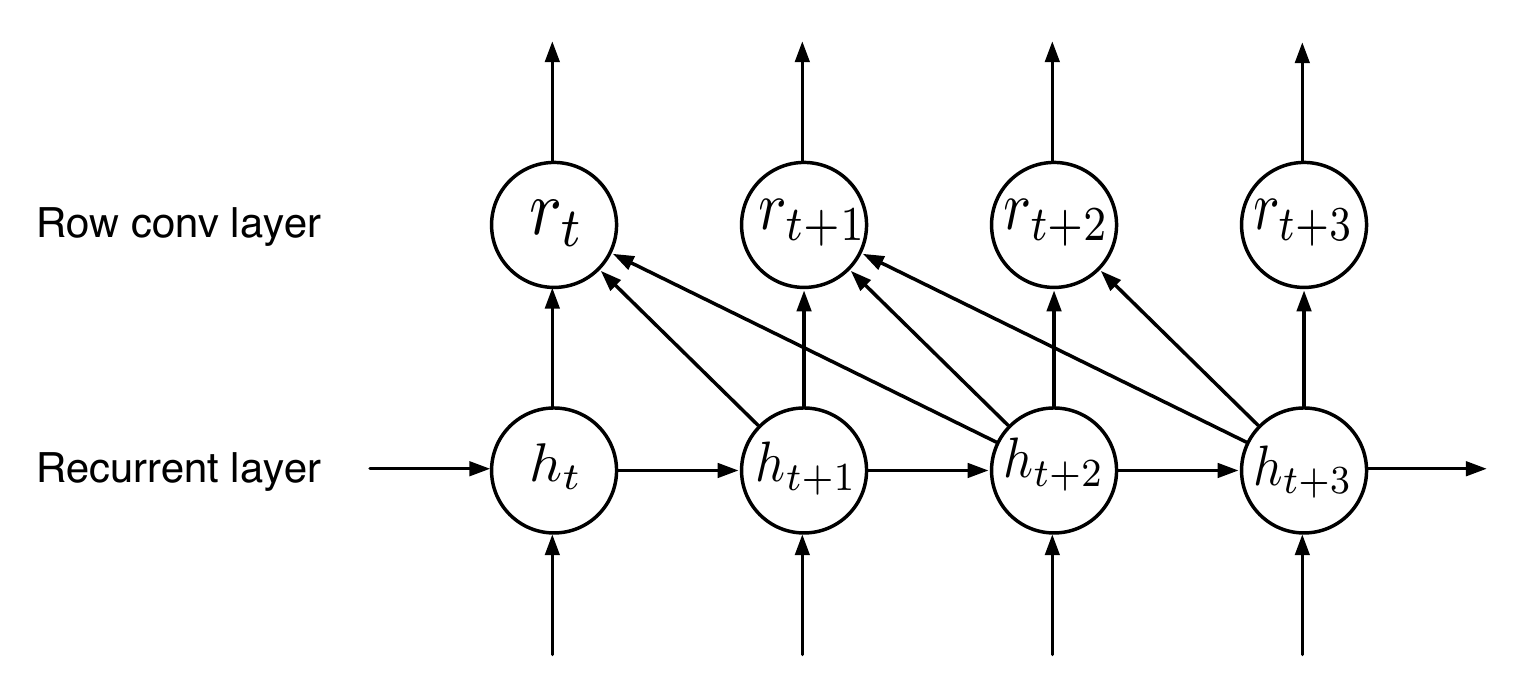}
\caption{Row convolution architecture with future context size of 2}
\label{fig:row-conv}
\end{figure}

To accomplish this, we employ a special layer that we call row convolution, shown in Figure~\ref{fig:row-conv}. The intuition behind this layer is that we only need a small portion of future information to make an accurate prediction at the current time-step. Suppose at time-step $t$, we use a future context of $\tau$ steps. We now have a feature matrix $h_{t:t+\tau} = [h_t, h_{t+1}, ..., h_{t+\tau}]$ of size $d\times(\tau+1)$. We define a parameter matrix $W$ of the same size as $h_{t:t+\tau}$. The activations $r_t$ for the new layer at time-step $t$ are 

\begin{align}
\label{eq:rowconv}
r_{t,i} = \sum_{j=1}^{\tau+1} W_{i,j} h_{t+j-1, i}, {\rm \ for\ } 1 \leq i \leq d.
\end{align}

Since the convolution-like operation in Eq.~\ref{eq:rowconv} is row oriented for both $W$ and $h_{t:t+\tau}$, we call this layer row convolution.

We place the row convolution layer above all recurrent layers. This has two advantages. First, this allows us to stream all computation below the row convolution layer on a finer granularity given little future context is needed. Second, this results in better CER compared to the best bidirectional model for Mandarin. We conjecture that the recurrent layers have learned good feature representations, so the row convolution layer simply gathers the appropriate information to feed to the classifier. Results for a unidirectional Mandarin speech system with row convolution and a comparison to a bidirectional model are given in Section~\ref{section:deployment} on deployment. 

\subsection{Language Model}
\label{subsection:languagemodel}

We train our RNN Models over millions of unique utterances, which enables the network to learn a powerful implicit language model. Our best models are quite adept at spelling, without any external language constraints. Further, in our development datasets we find many cases where our models can implicitly disambiguate homophones---for example, ``he expects the Japanese agent \emph{to} sell it for \emph{two} hundred seventy five thousand dollars''. Nevertheless, the labeled training data is small compared to the size of unlabeled text corpora that are available. Thus we find that WER improves when we supplement our system with a language model trained from external text. 

We use an \emph{n}-gram language model since they scale well to large amounts of unlabeled text~\cite{hannun2014deepspeech}. For English, our language model is a Kneser-Ney smoothed 5-gram model with pruning that is trained using the KenLM toolkit~\cite{heafield2013kenlm} on cleaned text from the Common Crawl Repository\footnote{\url{http://commoncrawl.org}}. The vocabulary is the most frequently used 400,000 words from 250 million lines of text, which produces a language model with about 850 million \emph{n}-grams. For Mandarin, the language model is a Kneser-Ney smoothed character level 5-gram model with pruning that is trained on an internal text corpus of 8 billion lines of text. This produces a language model with about 2 billion \emph{n}-grams.

During inference we search for the transcription $y$ that maximizes $Q(y)$ shown in Equation~\ref{eq:decoding}. This is a linear combination of $\log$ probabilities from the CTC trained network and language model, along with a word insertion term~\cite{hannun2014deepspeech}. 

\begin{equation}
\label{eq:decoding}
Q(y) = \log (p_{\textrm{ctc}}(y|x)) + \alpha \log(p_{\textrm{lm}}(y))  + \beta \: \textrm{word\_count}(y)
\end{equation}

The weight $\alpha$ controls the relative contributions of the language model and the CTC network. The weight $\beta$ encourages more words in the transcription. These parameters are tuned on a development set. We use a beam search to find the optimal transcription~\cite{hannun2014firstpass}.

\begin{table}
\centering
\begin{tabular}{l  l  r r r r r r}
\toprule
Language & Architecture & \multicolumn{3}{c}{Dev no LM} & \multicolumn{3}{c}{Dev LM} \\
\midrule
English  & 5-layer, 1 RNN & & 27.79 & & & 14.39 & \\
English  & 9-layer, 7 RNN & & 14.93 & & & 9.52 & \\
Mandarin & 5-layer, 1 RNN & & 9.80  & & & 7.13 & \\
Mandarin & 9-layer, 7 RNN & & 7.55  & & & 5.81 & \\
\bottomrule
\end{tabular}
\caption{Comparison of WER for English and CER for Mandarin with and without a language model. These are simple RNN models with only one layer of 1D invariant convolution.}
\label{table:languagemodels}
\end{table}

Table~\ref{table:languagemodels} shows that an external language model helps both English and Mandarin speech systems. The relative improvement given by the language model drops from 48\% to 36\% in English and 27\% to 23\% in Mandarin, as we go from a model with 5 layers and 1 recurrent layer to a model with 9 layers and 7 recurrent layers. We hypothesize that the network builds a stronger implicit language model with more recurrent layers. 

The relative performance improvement from a language model is higher in English than in Mandarin. We attribute this to the fact that a Chinese character represents a larger block of information than an English character. For example, if we output directly to syllables or words in English, the model would make fewer spelling mistakes and the language model would likely help less.

\subsection{Adaptation to Mandarin}
\label{subsection:chinesemodel}

The techniques that we have described so far can be used to build an end-to-end Mandarin speech recognition system that outputs Chinese characters directly. This precludes the need to construct a pronunciation model, which is often a fairly involved component for porting speech systems to other languages~\cite{shan2010}. Direct output to characters also precludes the need to explicitly model language specific pronunciation features. For example we do not need to model Mandarin tones explicitly, as some speech systems must do~\cite{shan2010, niu2013}.

The only architectural changes we make to our networks are due to the characteristics of the Chinese character set. Firstly, the output layer of the network outputs about 6000 characters, which includes the Roman alphabet, since hybrid Chinese-English transcripts are common. We incur an out of vocabulary error at evaluation time if a character is not contained in this set. This is not a major concern, as our test set has only 0.74\% out of vocab characters.

We use a character level language model in Mandarin as words are not usually segmented in text. The word insertion term of Equation~\ref{eq:decoding} becomes a character insertion term. In addition, we find that the performance of the beam search during decoding levels off at a smaller beam size. This allows us to use a beam size of 200 with a negligible degradation in CER. In Section~\ref{subsection:results-mandarin}, we show that our Mandarin speech models show roughly the same improvements to architectural changes as our English speech models.

\section{System Optimizations}
\label{section:optimization}

Our networks have tens of millions of parameters, and the training algorithm takes tens of single-precision exaFLOPs to converge. Since our ability to evaluate hypotheses about our data and models depends on the ability to train models quickly, we built a highly optimized training system. This system has two main components---a deep learning library written in \C++, along with a high-performance linear algebra library written in both CUDA and \C++. Our optimized software, running on dense compute nodes with 8 Titan X GPUs per node, allows us to sustain 24 single-precision teraFLOP/second when training a single model on one node. This is $45\%$ of the theoretical peak computational throughput of each node. We also can scale to multiple nodes, as outlined in the next subsection.

\subsection{Scalability and Data-Parallelism}
\label{subsection:scalability-and-data-parllelism}

We use the standard technique of data-parallelism to train on multiple GPUs using synchronous SGD. Our most common configuration uses a minibatch of $512$ on $8$ GPUs. Our training pipeline binds one process to each GPU. These processes then exchange gradient matrices during the backpropagation by using all-reduce, which exchanges a matrix between multiple processes and sums the result so that at the end, each process has a copy of the sum of all matrices from all processes. 

We find synchronous SGD useful because it is reproducible and deterministic. We have found that the appearance of non-determinism in our system often signals a serious bug, and so having reproducibility as a goal has greatly facilitates debugging. In contrast, asynchronous methods such as asynchronous SGD with parameter servers as found in Dean et al.~\cite{dean2012largescale} typically do not provide reproducibility and are therefore more difficult to debug. Synchronous SGD is simple to understand and implement. It scales well as we add multiple nodes to the training process.

\begin{figure}[h]
    \centering
    \includegraphics[width=0.6\textwidth]{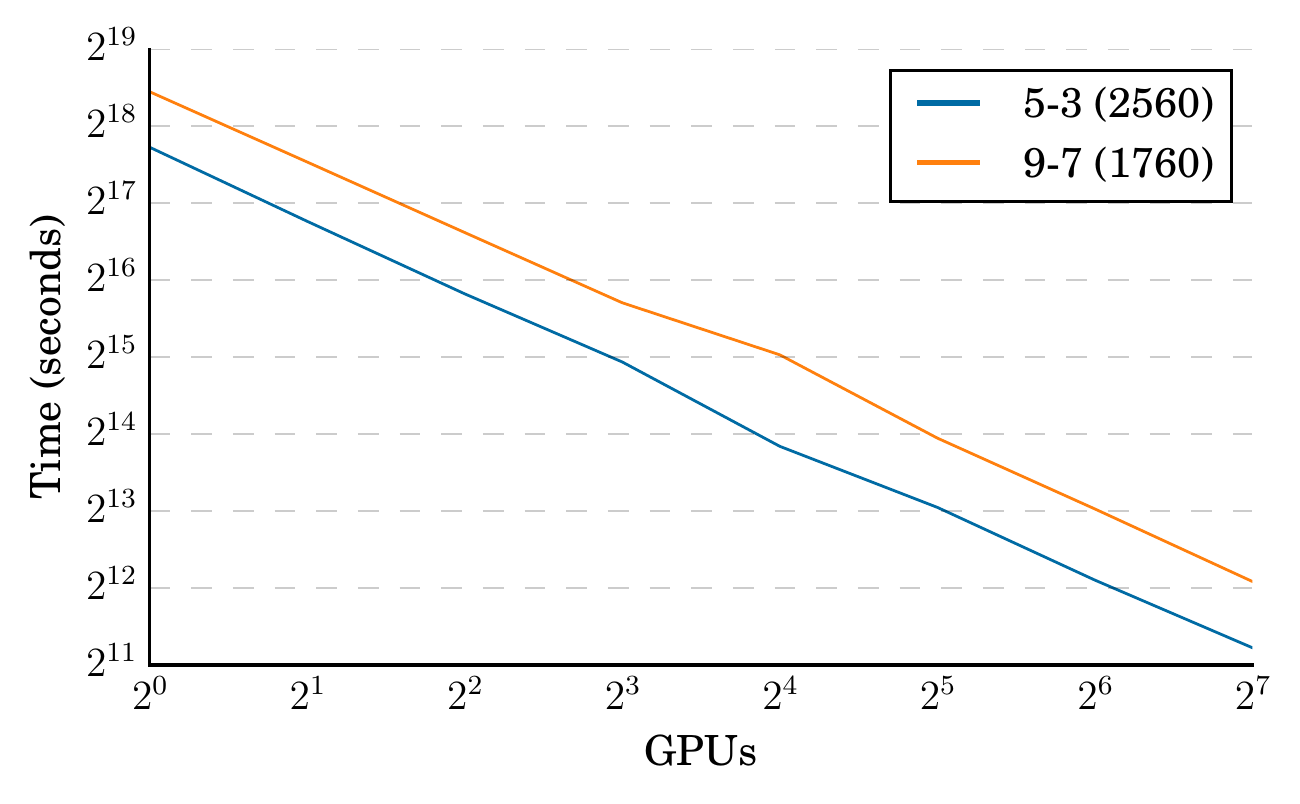}
    \caption{Scaling comparison of two networks---a 5 layer model with 3 recurrent layers containing 2560 hidden units in each layer and a 9 layer model with 7 recurrent layers containing 1760 hidden units in each layer. The times shown are to train 1 epoch. The 5 layer model trains faster because it uses larger matrices and is more computationally efficient.}
    \label{fig:speechdl-weakscaling}
\end{figure}

Figure~\ref{fig:speechdl-weakscaling} shows that time taken to train one epoch halves as we double the number of GPUs that we train on, thus achieving near-linear weak scaling. We keep the minibatch per GPU constant at 64 during this experiment, effectively doubling the minibatch as we double the number of GPUs. Although we have the ability to scale to large minibatches, we typically use either 8 or 16 GPUs during training with a minibatch of 512 or 1024, in order to converge to the best result.

Since all-reduce is critical to the scalability of our training, we wrote our own implementation of the ring algorithm~\cite{Patarasuk:2009:BOA:1482176.1482266, Thakur05optimizationof} for higher performance and better stability. Our implementation avoids extraneous copies between CPU and GPU, and is fundamental to our scalability. We configure OpenMPI with the \emph{smcuda} transport that can send and receive buffers residing in the memory of two different GPUs by using GPUDirect. When two GPUs are in the same PCI root complex, this avoids any unnecessary copies to CPU memory. This also takes advantage of tree-structured interconnects by running multiple segments of the ring concurrently between neighboring devices. We built our implementation using MPI send and receive, along with CUDA kernels for the element-wise operations. 

Table~\ref{table:compare-allreduce} compares the performance of our all-reduce implementation with that provided by OpenMPI version 1.8.5. We report the time spent in all-reduce for a full training run that ran for one epoch on our English dataset using a 5 layer, 3 recurrent layer architecture with $2560$ hidden units for all layers. In this table, we use a minibatch of 64 per GPU, expanding the algorithmic minibatch as we scale to more GPUs. We see that our implementation is considerably faster than OpenMPI's when the communication is within a node (8 GPUs or less). As we increase the number of GPUs and increase the amount of inter-node communication, the gap shrinks, although our implementation is still 2-4X faster. 

\begin{table}
\centering
\begin{tabular}{r  r r r  r r r  r r r}
\toprule
GPU & \multicolumn{3}{c}{OpenMPI} & \multicolumn{3}{c}{Our} & \multicolumn{3}{c}{Performance} \\
    & \multicolumn{3}{c}{all-reduce} & \multicolumn{3}{c}{all-reduce} & \multicolumn{3}{c}{Gain}        \\
\midrule
4   & & 55359.1 & & & 2587.4 & & & 21.4 &  \\
8   & & 48881.6 & & & 2470.9 & & & 19.8 &  \\
16  & & 21562.6 & & & 1393.7 & & & 15.5 &  \\
32  & & 8191.8  & & & 1339.6 & & & 6.1  &  \\
64  & & 1395.2  & & & 611.0  & & & 2.3  &  \\
128 & & 1602.1  & & & 422.6  & & & 3.8  &  \\
\bottomrule
\end{tabular}
\caption{Comparison of two different all-reduce implementations. All times are in seconds. Performance gain is the ratio of OpenMPI all-reduce time to our all-reduce time.}
\label{table:compare-allreduce}
\end{table}

All of our training runs use either 8 or 16 GPUs, and in this regime, our all-reduce implementation results in $2.5\times$ faster training for the full training run, compared to using OpenMPI directly. Optimizing all-reduce has thus resulted in important productivity benefits for our experiments, and has made our simple synchronous SGD approach scalable.

\subsection{GPU implementation of CTC loss function}

Calculating the CTC loss function is more complicated than performing forward and back propagation on our RNN architectures. Originally, we transferred activations from the GPUs to the CPU, where we calculated the loss function using an OpenMP parallelized implementation of CTC. However, this implementation limited our scalability rather significantly, for two reasons. Firstly, it became computationally more significant as we improved efficiency and scalability of the RNN itself. Secondly, transferring large activation matrices between CPU and GPU required us to spend interconnect bandwidth for CTC, rather than on transferring gradient matrices to allow us to scale using data parallelism to more processors.

To overcome this, we wrote a GPU implementation of the CTC loss function. Our parallel implementation relies on a slight refactoring to simplify the dependences in the CTC calculation, as well as the use of optimized parallel sort implementations from ModernGPU~\cite{ModernGPU}. We give more details of this parallelization in the Appendix.

\begin{table}
\centering
\begin{tabular}{l  l  c  c  c}
\toprule
Language  & Architecture   & CPU CTC Time & GPU CTC Time  & Speedup  \\
\midrule
English   & 5-layer, 3 RNN & 5888.12      & 203.56        & 28.9     \\
Mandarin  & 5-layer, 3 RNN & 1688.01      & 135.05        & 12.5     \\
\bottomrule
\end{tabular}
\caption{Comparison of time spent in seconds in computing the CTC loss function and gradient in one epoch for two different implementations. Speedup is the ratio of CPU CTC time to GPU CTC time.}
\label{table:gpucpuctc}
\end{table}

Table~\ref{table:gpucpuctc} compares the performance of two CTC implementations. The GPU implementation saves us 95 minutes per epoch in English, and 25 minutes in Mandarin. This reduces overall training time by 10-20\%, which is also an important productivity benefit for our experiments.

\subsection{Memory allocation}
Our system makes frequent use of dynamic memory allocations to GPU and CPU memory, mainly to store activation data for variable length utterances, and for intermediate results. Individual allocations can be very large; over 1 GB for the longest utterances.  For these very large allocations we found that CUDA's memory allocator and even \texttt{std::malloc} introduced significant overhead into our application---over a 2x slowdown from using \texttt{std::malloc} in some cases. This is because both \texttt{cudaMalloc} and \texttt{std::malloc} forward very large allocations to the operating system or GPU driver to update the system page tables. This is a good optimization for systems running multiple applications, all sharing memory resources, but editing page tables is pure overhead for our system where nodes are dedicated entirely to running a single model. To get around this limitation, we wrote our own memory allocator for both CPU and GPU allocations. Our implementation follows the approach of the last level shared allocator in jemalloc: all allocations are carved out of contiguous memory blocks using the buddy algorithm~\cite{Knowlton:1965:FSA:365628.365655}. To avoid fragmentation, we preallocate all of GPU memory at the start of training and subdivide individual allocations from this block. Similarly, we set the CPU memory block size that we forward to \texttt{mmap} to be substantially larger than \texttt{std::malloc}, at 12GB.

Most of the memory required for training deep recurrent networks is used to store activations through each layer for use by back propagation, not to store the parameters of the network. For example, storing the weights for a 70M parameter network with 9 layers requires approximately 280 MB of memory, but storing the activations for a batch of 64, seven-second utterances requires 1.5 GB of memory. TitanX GPUs include 12GB of GDDR5 RAM, and sometimes very deep networks can exceed the GPU memory capacity when processing long utterances. This can happen unpredictably, especially when the distribution of utterance lengths includes outliers, and it is desirable to avoid a catastrophic failure when this occurs. When a requested memory allocation exceeds available GPU memory, we allocate page-locked GPU-memory-mapped CPU memory using \texttt{cudaMallocHost} instead. This memory can be accessed directly by the GPU by forwarding individual memory transactions over PCIe at reduced bandwidth, and it allows a model to continue to make progress even after encountering an outlier.

The combination of fast memory allocation with a fallback mechanism that allows us to slightly overflow available GPU memory in exceptional cases makes the system significantly simpler, more robust, and more efficient.

\section{Training Data}
\label{section:data}

Large-scale deep learning systems require an abundance of labeled training data. We have collected an extensive training dataset for both English and Mandarin speech models, in addition to augmenting our training with publicly available datasets. In English we use 11,940 hours of labeled speech data containing 8 million utterances summarized in Table~\ref{table:englishdata}. For the Mandarin system we use 9,400 hours of labeled audio containing 11 million utterances. The Mandarin speech data consists of internal Baidu corpora, representing a mix of read speech and spontaneous speech, in both standard Mandarin and accented Mandarin.

\begin{table}
\centering
\begin{tabular}{l l r}
 \toprule
 Dataset & Speech Type & Hours \\
 \midrule
 WSJ          & read           &   80  \\
 Switchboard  & conversational &  300  \\
 Fisher       & conversational & 2000  \\
 LibriSpeech  & read           &  960  \\
 Baidu        & read           & 5000  \\
 Baidu        & mixed          & 3600  \\
 \midrule
 Total       &                &  11940 \\
 \bottomrule
\end{tabular}
\caption{Summary of the datasets used to train DS2 in English. The Wall Street Journal (WSJ), Switchboard and Fisher~\cite{cieri2004Fisher} corpora are all published by the Linguistic Data Consortium. The LibriSpeech dataset~\cite{panayotov2015} is available free on-line. The other datasets are internal Baidu corpora.}
\label{table:englishdata}
\end{table}

\subsection{Dataset Construction}
\label{subsection:dataconstruct}

Some of the internal English (3,600 hours) and Mandarin (1,400 hours) datasets were created from raw data captured as long audio clips with noisy transcriptions. The length of these clips ranged from several minutes to more than hour, making it impractical to unroll them in time in the RNN during training. To solve this problem, we developed an alignment, segmentation and filtering pipeline that can generate a training set with shorter utterances and few erroneous transcriptions.

The first step in the pipeline is to use an existing bidirectional RNN model trained with CTC to align the transcription to the frames of audio. For a given audio-transcript pair, $(x, y)$, we find the alignment that maximizes
\begin{equation}
\ell^* = \argmax_{\ell \in \text{Align}(x,y)} \prod_t^T p_{\text{ctc}}(\ell_t | x; \theta).
\end{equation}
This is essentially a Viterbi alignment found using a RNN model trained with CTC. Since Equation~\ref{eq:ctc} integrates over the alignment, the CTC loss function is never explicitly asked to produce an accurate alignment. In principle, CTC could choose to emit all the characters of the transcription after some fixed delay and this can happen with unidirectional RNNs~\cite{sak2015}. However, we found that CTC produces an accurate alignment when trained with a bidirectional RNN.

Following the alignment is a segmentation step that splices the audio and the corresponding aligned transcription whenever it encounters a long series of consecutive \emph{blank} labels occurs, since this usually denotes a stretch of silence. By tuning the number of consecutive \emph{blank}s, we can tune the length of the utterances generated. For the English speech data, we also require a \emph{space} token to be within the stretch of \emph{blank}s in order to segment only on word boundaries. We tune the segmentation to generate utterances that are on average 7 seconds long.

The final step in the pipeline removes erroneous examples that arise from a failed alignment. We crowd source the ground truth transcriptions for several thousand examples. The word level edit distance between the ground truth and the aligned transcription is used to produce a \emph{good} or \emph{bad} label. The threshold for the word level edit distance is chosen such that the resulting WER of the \emph{good} portion of the development set is less than 5\%. We then train a linear classifier to accurately predict bad examples given the input features generated from the speech recognizer. We find the following features useful: the raw CTC cost, the CTC cost normalized by the sequence length, the CTC cost normalized by the transcript length, the ratio of the sequence length to the transcript length, the number of words in the transcription and the number of characters in the transcription. For the English dataset, we find that the filtering pipeline reduces the WER from 17\% to 5\% while retaining more than 50\% of the examples.

\subsection{Data Augmentation}
\label{subsection:noisesynth}

We augment our training data by adding noise to increase the effective size of our training data and to improve our robustness to noisy speech~\cite{hannun2014deepspeech}. Although the training data contains some intrinsic noise, we can increase the quantity and variety of noise through augmentation. Too much noise augmentation tends to make optimization difficult and can lead to worse results, and too little noise augmentation makes the system less robust to low signal-to-noise speech.  We find that a good balance is to add noise to 40\% of the utterances that are chosen at random. The noise source consists of several thousand hours of randomly selected audio clips combined to produce hundreds of hours of noise.

\subsection{Scaling Data}

Our English and Mandarin corpora are substantially larger than those commonly reported in speech recognition literature. In Table~\ref{table:datascale}, we show the effect of increasing the amount of labeled training data on WER. This is done by randomly sampling the full dataset before training. For each dataset, the model was trained for up to 20 epochs though usually early-stopped based on the error on a held out development set. We note that the WER decreases with a power law for both the regular and noisy development sets. The WER decreases by $\sim$40\% relative for each factor of 10 increase in training set size. We also observe a consistent gap in WER ($\sim$60\% relative) between the regular and noisy datasets, implying that more data benefits both cases equally. 

This implies that a speech system will continue to improve with more labeled training data. We hypothesize that equally as important as increasing raw number of hours is increasing the number of speech \emph{contexts} that are captured in the dataset. A context can be any property that makes speech unique including different speakers, background noise, environment, and microphone hardware. While we do not have the labels needed to validate this claim, we suspect that measuring WER as a function of speakers in the dataset would lead to much larger relative gains than simple random sampling.

\begin{table}
\centering
\begin{tabular}{r r r  r  r r r  r r r}
\toprule
\multicolumn{3}{c}{Fraction of Data} & Hours & \multicolumn{3}{c}{Regular Dev} & \multicolumn{3}{c}{Noisy Dev} \\
\midrule
& 1\%   & & 120   & & 29.23 & & & 50.97 & \\
& 10\%  & & 1200  & & 13.80 & & & 22.99 & \\
& 20\%  & & 2400  & & 11.65 & & & 20.41 & \\
& 50\%  & & 6000  & & 9.51  & & & 15.90 & \\
& 100\% & & 12000 & & 8.46  & & & 13.59 & \\
\bottomrule
\end{tabular}
\caption{Comparison of English WER for Regular and Noisy development sets on increasing training dataset size. The architecture is a 9-layer model with 2 layers of 2D-invariant convolution and 7 recurrent layers with 68M parameters.}
\label{table:datascale}
\end{table}

\section{Results}
\label{section:results}

To better assess the real-world applicability of our speech system, we evaluate on a wide range of test sets. We use several publicly available benchmarks and several test sets collected internally. Together these test sets represent a wide range of challenging speech environments including low signal-to-noise ratios (noisy and far-field), accented, read, spontaneous and conversational speech. 

All models are trained for 20 epochs on either the full English dataset, described in Table~\ref{table:englishdata}, or the full Mandarin dataset described in Section~\ref{section:data}. We use stochastic gradient descent with Nesterov momentum~\cite{sutskever2013nag} along with a minibatch of 512 utterances. If the norm of the gradient exceeds a threshold of 400, it is rescaled to 400~\cite{pascanu2012}. The model which performs the best on a held-out development set during training is chosen for evaluation. The learning rate is chosen from $[1\times10^{-4}, 6\times10^{-4}]$ to yield fastest convergence and annealed by a constant factor of 1.2 after each epoch. We use a momentum of 0.99 for all models.

The language models used are those described in Section~\ref{subsection:languagemodel}. The decoding parameters from Equation~\ref{eq:decoding} are tuned on a held-out development set. We use a beam size of 500 for the English decoder and a beam size of 200 for the Mandarin decoder.

\subsection{English}
\label{subsection:english-results}

The best DS2 model has 11 layers with 3 layers of 2D convolution, 7 bidirectional recurrent layers, a fully-connected output layer along with Batch Normalization. The first layer outputs to bigrams with a temporal stride of 3. By comparison the DS1 model has 5 layers with a single bidirectional recurrent layer and it outputs to unigrams with a temporal stride of 2 in the first layer. We report results on several test sets for both the DS2 and DS1 model. We do not tune or adapt either model to any of the speech conditions in the test sets. Language model decoding parameters are set once on a held-out development set.

To put the performance of our system in context, we benchmark most of our results against human workers, since speech recognition is an audio perception and language understanding problem that humans excel at. We obtain a measure of human level performance by paying workers from Amazon Mechanical Turk to hand-transcribe all of our test sets. Two workers transcribe the same audio clip, that is typically about 5 seconds long, and we use the better of the two transcriptions for the final WER calculation. They are free to listen to the audio clip as many times as they like. These workers are mostly based in the United States, and on average spend about 27 seconds per transcription. The hand-transcribed results are compared to the existing ground truth to produce a WER. While the existing ground truth transcriptions do have some label error, this is rarely more than 1\%. This implies that disagreement between the ground truth transcripts and the human level transcripts is a good heuristic for human level performance.

\subsubsection{Model Size}

Our English speech training set is substantially larger than the size of commonly used speech datasets. Furthermore, the data is augmented with noise synthesis. To get the best generalization error, we expect that the model size must increase to fully exploit the patterns in the data. In Section~\ref{subsection:depth} we explored the effect of model depth while fixing the number of parameters. In contrast, here we show the effect of varying model size on the performance of the speech system. We only vary the size of each layer, while keeping the depth and other architectural parameters constant. We evaluate the models on the same Regular and Noisy development sets that we use in Section~\ref{subsection:2dconv}.

\begin{table}
\centering
\begin{tabular}{r  c  r r r  r r r}
\toprule
Model size & Model type & \multicolumn{3}{c}{Regular Dev} & \multicolumn{3}{c}{Noisy Dev} \\
\midrule
$18 \times 10^6$     & GRU &   & 10.59 & &  & 21.38 & \\
$38 \times 10^6$     & GRU &   & 9.06  & &  & 17.07 & \\
$70 \times 10^6$     & GRU &   & 8.54  & &  & 15.98 & \\
$70 \times 10^6$     & RNN &   & 8.44  & &  & 15.09 & \\
$100 \times 10^6$    & GRU &   & 7.78  & &  & 14.17 & \\
$100 \times 10^6$    & RNN &   & 7.73  & &  & 13.06 & \\
\bottomrule
\end{tabular}
\caption{Comparing the effect of model size on the WER of the English speech system on both the regular and noisy development sets. We vary the number of hidden units in all but the convolutional layers. The GRU model has 3 layers of bidirectional GRUs with 1 layer of 2D-invariant convolution. The RNN model has 7 layers of bidirectional simple recurrence with 3 layers of 2D-invariant convolution. Both models output bigrams with a temporal stride of 3. All models contain approximately 35 million parameters and are trained with BatchNorm and SortaGrad.}
\label{table:modelsize}
\end{table}

The models in Table~\ref{table:modelsize} differ from those in Table~\ref{table:rnns} in that we increase the the stride to 3 and output to bigrams. Because we increase the model size to as many as 100 million parameters, we find that an increase in stride is necessary for fast computation and memory constraints. However, in this regime we note that the performance advantage of the GRU networks appears to diminish over the simple RNN. In fact, for the 100 million parameter networks the simple RNN performs better than the GRU network and is faster to train despite the 2 extra layers of convolution.

Table~\ref{table:modelsize} shows that the performance of the system improves consistently up to 100 million parameters. All further English DS2 results are reported with this same 100 million parameter RNN model since it achieves the lowest generalization errors.

Table~\ref{table:test} shows that the 100 million parameter RNN model (DS2) gives a 43.4\% relative improvement over the 5-layer model with 1 recurrent layer (DS1) on an internal Baidu dataset of 3,300 utterances that contains a wide variety of speech including challenging accents, low signal-to-noise ratios from far-field or background noise, spontaneous and conversational speech. 

\begin{table}
\centering
\begin{tabular}{l  c  c  c}
\toprule
Test set   & DS1 & DS2 \\
\midrule
Baidu Test & 24.01  & 13.59 \\
\bottomrule
\end{tabular}
\caption{Comparison of DS1 and DS2 WER on an internal test set of 3,300 examples. The test set contains a wide variety of speech including accents, low signal-to-noise speech, spontaneous and conversational speech.}
\label{table:test}
\end{table}

\subsubsection{Read Speech}

Read speech with high signal-to-noise ratio is arguably the easiest large vocabulary for a continuous speech recognition task. We benchmark our system on two test sets from the Wall Street Journal (WSJ) corpus of read news articles. These are available in the LDC catalog as LDC94S13B and LDC93S6B. We also take advantage of the recently developed LibriSpeech corpus constructed using audio books from the LibriVox project~\cite{panayotov2015}.

Table~\ref{table:readspeech} shows that the DS2 system outperforms humans in 3 out of the 4 test sets and is competitive on the fourth. Given this result, we suspect that there is little room for a generic speech system to further improve on clean read speech without further domain adaptation.

\begin{table}
\centering
\begin{tabular}{l  r  r r}
\toprule
\multicolumn{4}{c}{Read Speech}\\
\midrule
Test set               & DS1   & DS2 &  Human \\ 
\midrule
WSJ eval'92            & 4.94  & 3.60  & 5.03 \\ 
WSJ eval'93            & 6.94  & 4.98  & 8.08 \\ 
LibriSpeech test-clean & 7.89  & 5.33  & 5.83 \\ 
LibriSpeech test-other & 21.74 & 13.25 & 12.69 \\ 
\bottomrule
\end{tabular}
\caption{Comparison of WER for two speech systems and human level performance on read speech.}
\label{table:readspeech}
\end{table}

\subsubsection{Accented Speech}

Our source for accented speech is the publicly available VoxForge (\url{http://www.voxforge.org}) dataset, which has clean speech read from speakers with many different accents. We group these accents into four categories. The American-Canadian and Indian groups are self-explanatory. The Commonwealth accent denotes speakers with British, Irish, South African, Australian and New Zealand accents. The European group contains speakers with accents from countries in Europe that do not have English as a first language. We construct a test set from the VoxForge data with 1024 examples from each accent group for a total of 4096 examples.

Performance on these test sets is to some extent a measure of the breadth and quality of our training data. Table~\ref{table:voxforge} shows that our performance improved on all the accents when we include more accented training data and use an architecture that can effectively train on that data. However human level performance is still notably better than that of DS2 for all but the Indian accent. 

\begin{table}
\centering
\begin{tabular}{l  r  r  r}
\toprule
\multicolumn{4}{c}{Accented Speech}\\
\midrule
Test set                   & DS1   & DS2 & Human \\
\midrule
VoxForge American-Canadian & 15.01 & 7.55  & 4.85 \\
VoxForge Commonwealth      & 28.46 & 13.56 & 8.15 \\
VoxForge European          & 31.20 & 17.55 & 12.76 \\
VoxForge Indian            & 45.35 & 22.44 & 22.15 \\
\bottomrule
\end{tabular}
\caption{Comparing WER of the DS1 system to the DS2 system on accented speech.}
\label{table:voxforge}
\end{table}

\subsubsection{Noisy Speech}

We test our performance on noisy speech using the publicly available test sets from the recently completed third CHiME challenge~\cite{barker2015chime}. This dataset has 1320 utterances from the WSJ test set read in various noisy environments, including a bus, a cafe, a street and a pedestrian area. The CHiME set also includes 1320 utterances with simulated noise from the same environments as well as the control set containing the same utterances delivered by the same speakers in a noise-free environment. Differences between results on the control set and the noisy sets provide a measure of the network's ability to handle a variety of real and synthetic noise conditions. The CHiME audio has 6 channels and using all of them can provide substantial performance improvements~\cite{yoshioka2015}. We use a {\it single} channel for all our results, since multi-channel audio is not pervasive on most devices. Table~\ref{table:chime} shows that DS2 substantially improves upon DS1, however DS2 is worse than human level performance on noisy data. The relative gap between DS2 and human level performance is larger when the data comes from a real noisy environment instead of synthetically adding noise to clean speech.

\begin{table}
\centering
\begin{tabular}{l  r  r r}
\toprule
\multicolumn{4}{c}{Noisy Speech}\\
\midrule
Test set & DS1 & DS2  &  Human \\
\midrule
CHiME eval clean & 6.30  & 3.34  & 3.46 \\
CHiME eval real  & 67.94 & 21.79 & 11.84 \\
CHiME eval sim   & 80.27 & 45.05 & 31.33 \\
\bottomrule
\end{tabular}
\caption{Comparison of DS1 and DS2 system on noisy speech. ``CHiME eval clean'' is a noise-free baseline. The ``CHiME eval real'' dataset is collected in real noisy environments and the ``CHiME eval sim'' dataset has similar noise synthetically added to clean speech. Note that we use only one of the six channels to test each utterance.}
\label{table:chime}
\end{table}

\subsection{Mandarin}
\label{subsection:results-mandarin}

In Table~\ref{table:results-mandarin} we compare several architectures trained on the Mandarin Chinese speech, on a development set of 2000 utterances as well as a test set of 1882 examples of noisy speech. This development set was also used to tune the decoding parameters
We see that the deepest model with 2D-invariant convolution and BatchNorm outperforms the shallow RNN by 48\% relative, thus continuing the trend that we saw with the English system---multiple layers of bidirectional recurrence improves performance substantially. 

\begin{table}[ht!]
\centering
\begin{tabular}{l  r  r  }
\toprule
Architecture & Dev & Test \\
\midrule
5-layer, 1 RNN                & 7.13  & 15.41 \\
5-layer, 3 RNN                & 6.49  & 11.85 \\
5-layer, 3 RNN + BatchNorm           & 6.22  & 9.39 \\
9-layer, 7 RNN + BatchNorm + 2D conv & 5.81  & 7.93 \\
\bottomrule
\end{tabular}
\caption{Comparison of the improvements in DeepSpeech with architectural improvements. The development and test sets are Baidu internal corpora. All the models in the table have about 80 million parameters each}
\label{table:results-mandarin}
\end{table}

We find that our best Mandarin Chinese speech system transcribes short voice-query like utterances better than a typical Mandarin Chinese speaker. To benchmark against humans we ran a test with 100 randomly selected utterances and had a group of 5 humans label all of them together. The group of humans had an error rate of 4.0\% as compared to the speech systems performance of 3.7\%. We also compared a single human transcriber to the speech system on 250 randomly selected utterances. In this case the speech system performs much better: 9.7\% for the human compared to 5.7\% for the speech model.

\section{Deployment}
\label{section:deployment}

Real-world applications usually require a speech system to transcribe in real time or with relatively low latency. The system used in Section~\ref{subsection:english-results} is not well-designed for this task, for several reasons. First, since the RNN has several bidirectional layers, transcribing the first part of an utterance requires the entire utterance to be presented to the RNN. Second, since we use a wide beam when decoding with a language model, beam search can be expensive, particularly in Mandarin where the number of possible next characters is very large (around 6000). Third, as described in Section~\ref{section:model}, we normalize power across an entire utterance, which again requires the entire utterance to be available in advance.

We solve the power normalization problem by using some statistics from our training set to perform an adaptive normalization of speech inputs during online transcription. We can solve the other problems by modifying our network and decoding procedure to produce a model that performs almost as well while having much lower latency. We focus on our Mandarin system since some aspects of that system are more challenging to deploy (e.g. the large character set), but the same techniques could also be applied in English.

In this section, latency refers to the computational latency of our speech system as measured from the end of an utterance until the transcription is produced. This latency does not include data transmission over the internet, and does not measure latency from the beginning of an utterance until the first transcription is produced. We focus on latency from end of utterance to transcription because it is important to applications using speech recognition.

\subsection{Batch Dispatch}
\label{subsection:batching}

\begin{figure}
\centering
\includegraphics[width=0.6\textwidth]{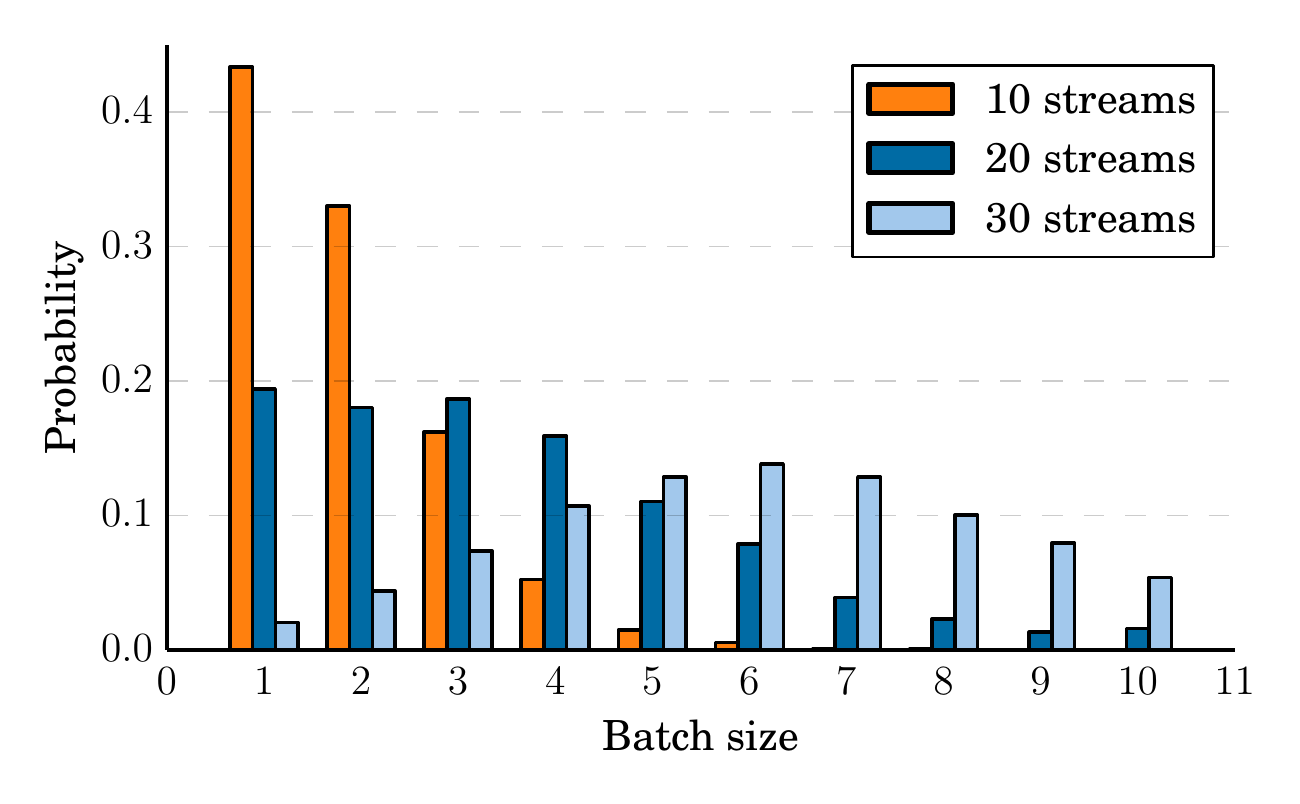}
\caption{Probability that a request is processed in a batch of given size}
\label{fig:batching}
\end{figure}

In order to deploy our relatively large deep neural networks at low latency, we have paid special attention to efficiency during deployment. Most internet applications process requests individually as they arrive in the data center. This makes for a straightforward implementation where each request can be managed by one thread. However, processing requests individually is inefficient computationally, for two main reasons. Firstly, when processing requests individually, the processor must load all the weights of the network for each request. This lowers the arithmetic intensity of the workload, and tends to make the computation memory bandwidth bound, as it is difficult to effectively use on-chip caches when requests are presented individually. Secondly, the amount of parallelism that can be exploited to classify one request is limited, making it difficult to exploit SIMD or multi-core parallelism. RNNs are especially challenging to deploy because evaluating RNNs sample by sample relies on sequential matrix vector multiplications, which are bandwidth bound and difficult to parallelize.

To overcome these issues, we built a batching scheduler called Batch Dispatch that assembles streams of data from user requests into batches before performing forward propagation on these batches. In this case, there is a tradeoff between increased batch size, and consequently improved efficiency, and increased latency. The more we buffer user requests to assemble a large batch, the longer users must wait for their results. This places constraints on the amount of batching we can perform.

We use an eager batching scheme that processes each batch as soon as the previous batch is completed, regardless of how much work is ready by that point. This scheduling algorithm has proved to be the best at reducing end-user latency, despite the fact that it is less efficient computationally, since it does not attempt to maximize batch size.

Figure~\ref{fig:batching} shows the probability that a request is processed in a batch of given size for our production system running on a single NVIDIA Quadro K1200 GPU, with 10-30 concurrent user requests. As expected, batching works best when the server is heavily loaded: as load increases, the distribution shifts to favor processing requests in larger batches. However, even with a light load of only 10 concurrent user requests, our system performs more than half the work in batches with at least 2 samples.

\begin{figure}[h]
\centering
\includegraphics[width=0.6\textwidth]{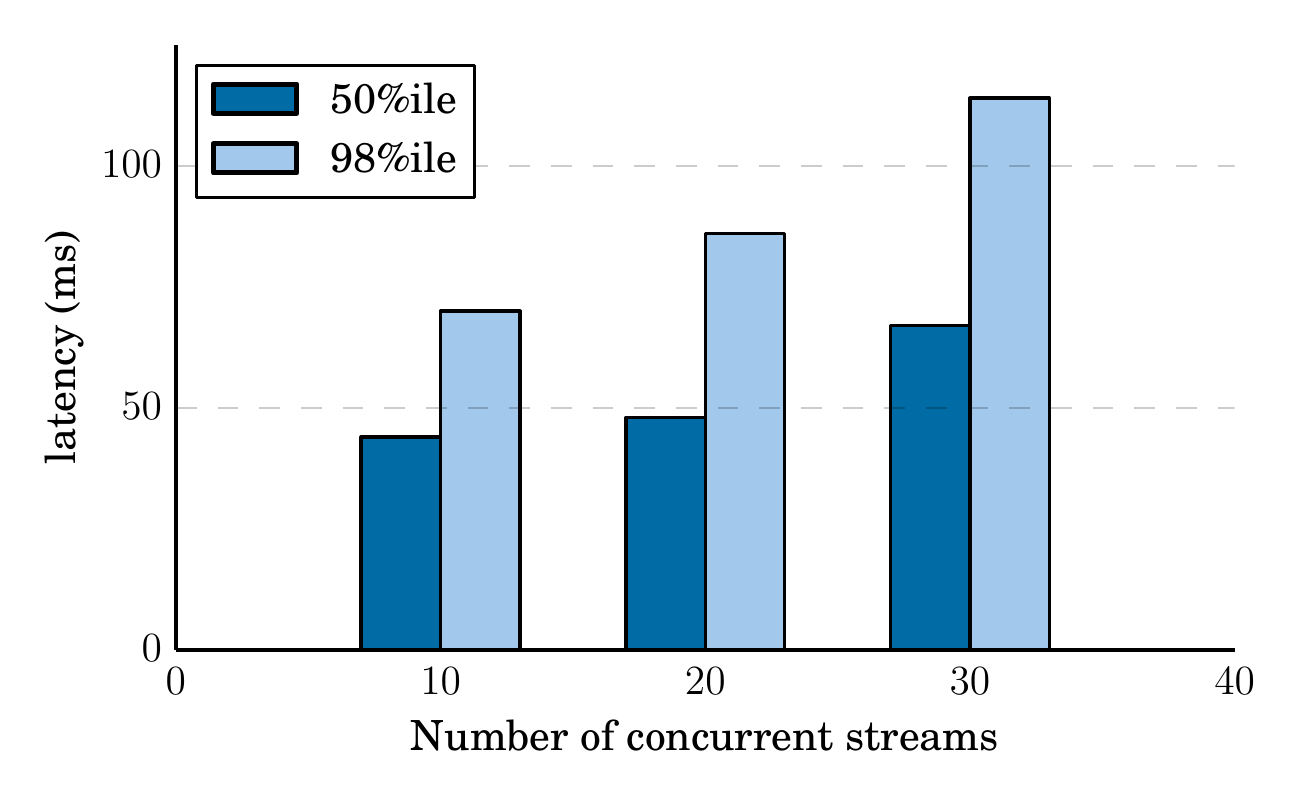}
\caption{Median and 98 percentile latencies as a function of server load}
\label{fig:latency}
\end{figure}

We see in Figure~\ref{fig:latency}, that our system achieves a median latency of 44 ms, and a 98 percentile latency of 70 ms when loaded with 10 concurrent streams. As the load increases on the server, the batching scheduler shifts work to more efficient batches, which keeps latency low. This shows that Batch Dispatch makes it possible to deploy these large models at high throughput and low latency.

\subsection{Deployment Optimized Matrix Multiply Kernels}
We have found that deploying our models using half-precision (16-bit) floating-point arithmetic does not measurably change recognition accuracy. Because deployment does not require any updates to the network weights, it is far less sensitive to numerical precision than training. Using half-precision arithmetic saves memory space and bandwidth, which is especially useful for deployment, since RNN evaluation is dominated by the cost of caching and streaming the weight matrices.

As seen in Section~\ref{subsection:batching}, the batch size during deployment is much smaller than in training. We found that standard BLAS libraries are inefficient at this batch size. To overcome this, we wrote our own half-precision matrix-matrix multiply kernel. For 10 simultaneous streams over 90 percent of batches are for $N \leq 4$, a regime where the matrix multiply will be bandwidth bound.  We store the $A$ matrix transposed to maximize bandwidth by using the widest possible vector loads while avoiding transposition after loading.  Each warp computes four rows of output for all $N$ output columns.  Note that for $N \leq 4$ the $B$ matrix fits entirely in the L1 cache.  This scheme achieves 90 percent of peak bandwidth for $N \leq 4$ but starts to lose efficiency for larger $N$ as the $B$ matrix stops fitting into the L1 cache.  Nonetheless, it continues to provide improved performance over existing libraries up to $N=10$.

\begin{figure}
\centering
\includegraphics[width=0.6\textwidth]{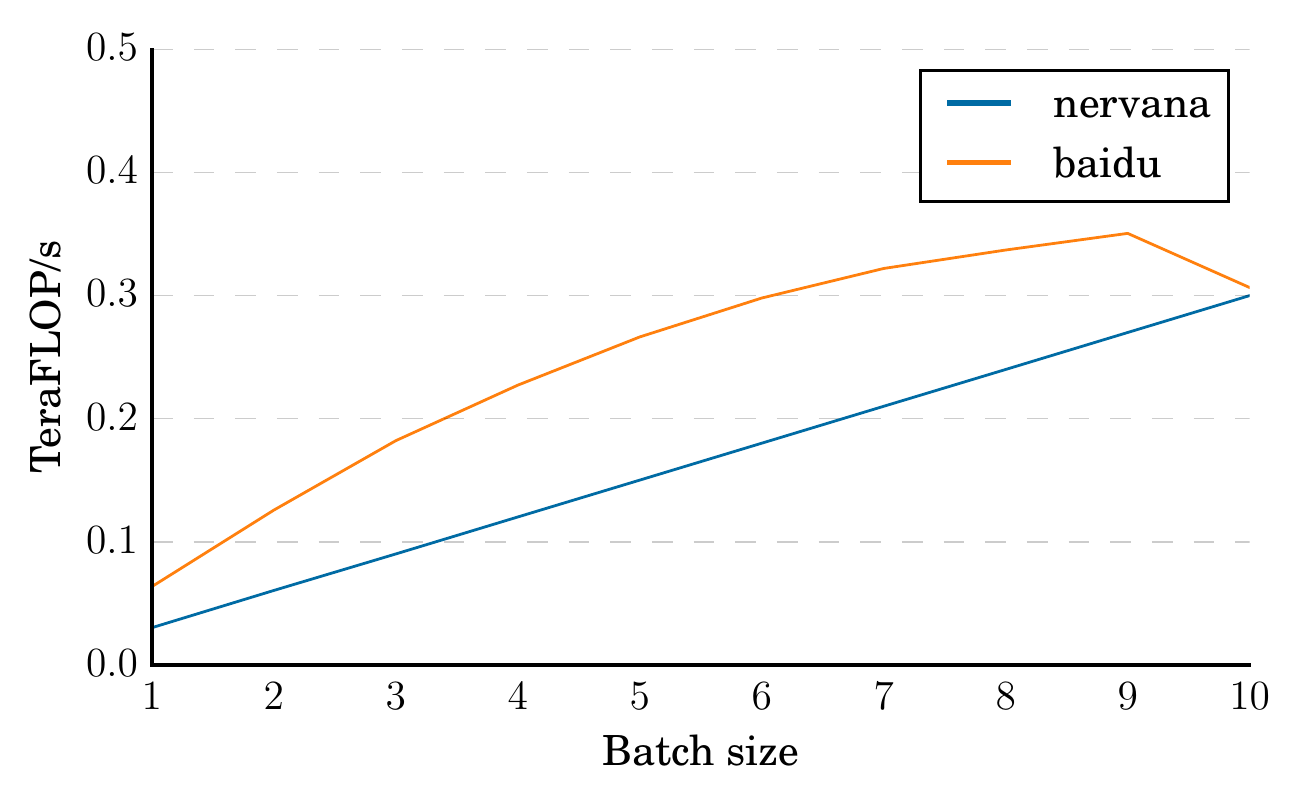}
\caption{Comparison of kernels that compute $A x = b$ where $A$ is a matrix with dimension $2560 \times 2560$, and $x$ is a matrix with dimension $2560 \times \textrm{Batch size}$, where $\textrm{Batch size} \in [1, 10]$. All matrices are in half-precision format.}
\label{fig:HGEMM}
\end{figure}

Figure~\ref{fig:HGEMM} shows that our deployment kernel sustains a higher computational throughput than those from Nervana Systems~\cite{NervanaGPU} on the K1200 GPU, across the entire range of batch sizes that we use in deployment. Both our kernels and the Nervana kernels are significantly faster than NVIDIA CUBLAS version 7.0, more details are found here~\cite{Elsen2015}.

\subsection{Beam Search}

Performing the beam search involves repeated lookups in the \emph{n}-gram language model, most of which translate to uncached reads from memory. The direct implementation of beam search means that each time-step dispatches one lookup per character for each beam. In Mandarin, this results in over 1M lookups per 40ms stride of speech data, which is too slow for deployment. To deal with this problem, we use a heuristic to further prune the beam search. Rather than considering all characters as viable additions to the beam, we only consider the fewest number of characters whose cumulative probability is at least $p$. In practice, we have found that $p=0.99$ works well. Additionally, we limit ourselves to no more than 40 characters. This speeds up the Mandarin language model lookup time by a factor of 150x, and has a negligible effect on the CER (0.1-0.3\% relative).

\subsection{Results}

We can deploy our system at low latency and high throughput without sacrificing much accuracy. On a held-out set of 2000 utterances, our research system achieves 5.81 character error rate whereas the deployed system achieves 6.10 character error rate. This is only a 5\% relative degradation for the deployed system. In order to accomplish this, we employ a neural network architecture with low deployment latency, reduce the precision of our network to 16-bit, built a batching scheduler to more efficiently evaluate RNNs, and find a simple heuristic to reduce beam search cost. The model has five forward-only recurrent layers with 2560 hidden units, one row convolution layer (Section~\ref{section:fom}) with $\tau=19$, and one fully-connected layer with 2560 hidden units. These techniques allow us to deploy Deep Speech at low cost to interactive applications.

\section{Conclusion}
End-to-end deep learning presents the exciting opportunity to improve speech recognition systems continually with increases in data and computation.  Indeed, our results show that, compared to the previous incarnation, Deep Speech has significantly closed the gap in transcription performance with human workers by leveraging more data and larger models.  Further, since the approach is highly generic, we've shown that it can quickly be applied to new languages. Creating high-performing recognizers for two very different languages, English and Mandarin, required essentially no expert knowledge of the languages.  Finally, we have also shown that this approach can be efficiently deployed by batching user requests together on a GPU server, paving the way to deliver end-to-end Deep Learning technologies to users. 

To achieve these results, we have explored various network architectures, finding several effective techniques:  enhancements to numerical optimization through SortaGrad and Batch Normalization, evaluation of RNNs with larger strides with bigram outputs for English, searching through both bidirectional and unidirectional models. This exploration was powered by a well optimized, High Performance Computing inspired training system that allows us to train new, full-scale models on our large datasets in just a few days.

Overall, we believe our results confirm and exemplify the value of end-to-end Deep Learning methods for speech recognition in several settings.  In those cases where our system is not already comparable to humans, the difference has fallen rapidly, largely because of application-agnostic Deep Learning techniques.  We believe these techniques will continue to scale, and thus conclude that the vision of a single speech system that outperforms humans in most scenarios is imminently achievable.

\section*{Acknowledgments} 
We are grateful to Baidu's speech technology group for help with data preparation and useful conversations. We would like to thank Scott Gray, Amir Khosrowshahi and all of Nervana Systems for their excellent matrix multiply routines and useful discussions.  We would also like to thank Natalia Gimelshein of NVIDIA  for useful discussions and thoughts on implementing our fast deployment matrix multiply.

{\small
\bibliography{references}
\bibliographystyle{abbrv}
}

\appendix
\section{Scalability improvements}
In this section, we discuss some of our scalability improvements in more detail.

\subsection{Node and cluster architecture}
The software stack runs on a compute dense node built from $2$ Intel CPUs and $8$ NVIDIA Titan X GPUs, with peak single-precision computational throughput of $53$ teraFLOP/second. Each node also has $384$ GB of CPU memory and an $8$ TB storage volume built from two 4 TB hard disks in RAID-0 configuration. We use the CPU memory to cache our input data so that we are not directly exposed to the low bandwidth and high latency of spinning disks. We replicate our English and Mandarin datasets on each node's local hard disk. This allows us to use our network only for weight updates and avoids having to rely on centralized file servers.

\begin{figure}[h]
    \centering
    \includegraphics[width=0.8\textwidth]{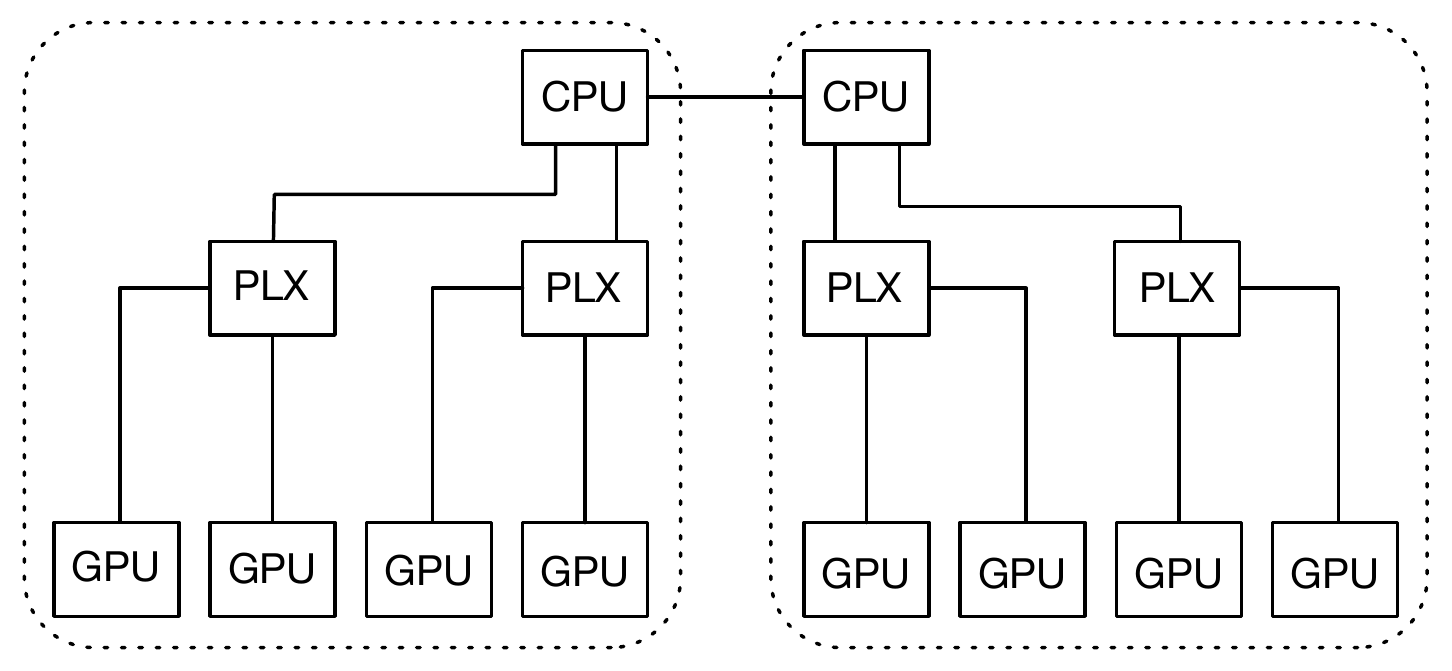}
    \caption{Schematic of our training node where PLX indicates a PCI switch and the dotted box includes all devices that are connected by the same PCI root complex.}
    \label{fig:node-arch}
\end{figure}

Figure~\ref{fig:node-arch} shows a schematic diagram of one our nodes, where all devices connected by the same PCI root complex are encapsulated in a dotted box. We have tried to maximize the number of GPUs within the root complex for faster communication between GPUs using GPUDirect. This allows us to use an efficient communication mechanism to transfer gradient matrices between GPUs. 

All the nodes in our cluster are connected through Fourteen Data Rate (FDR) Infiniband which is primarily used for gradient transfer during back-propagation.

\subsection{GPU Implementation of CTC Loss Function}
\label{appendix:GPUCTC}

The CTC loss function that we use to train our models has two passes: forward and backward, and the gradient computation involves element-wise addition of two matrices, $\alpha$ and $\beta$, generated during the forward and backward passes respectively. Finally, we sum the gradients using the character in the utterance label as the key, to generate one gradient per character. These gradients are then back-propagated through the network. The input to the CTC loss function are probabilities calculated by the softmax function which can be very small, so we compute in $\log$ probability space for better numerical stability.

The forward pass of the CTC algorithm calculates the $\alpha$ matrix, which has $\mathsf{S}$ rows and $\mathsf{T}$ columns, where $\mathsf{S} = 2 (\mathsf{L} + 1)$. The variable $\mathsf{L}$ is the number of characters in the label and $\mathsf{T}$ is the number of time-steps in the utterance. Our CPU-based implementation of the CTC algorithm assigns one thread to each utterance label in a minibatch, performing the CTC calculation for the utterances in parallel. Each thread calculates the relevant entries of the matrix sequentially. This is inefficient for two reasons. 

Firstly, since the remainder of our network is computed on the GPU, the output of the softmax function has to be copied to the CPU for CTC calculation. The gradient matrices from the CTC function then has to be copied back to the GPU for backpropagation. For languages like Mandarin with large character sets, these matrices have hundreds of millions of entries, making this copy expensive. Furthermore, we need as much interconnect bandwidth as possible for synchronizing the gradient updates with data parallelism, so this copy incurs a substantial opportunity cost.

Secondly, although entries in each column of the $\alpha$ matrix can be computed in parallel, the number of entries to calculate in each column depends both on the column and the number of repeated characters in the utterance label. Due to this complexity, the CPU implementation does not use SIMD parallelism optimally, making the computation inefficient.

\begin{figure}[h]
    \centering
    \includegraphics[width=0.6\textwidth]{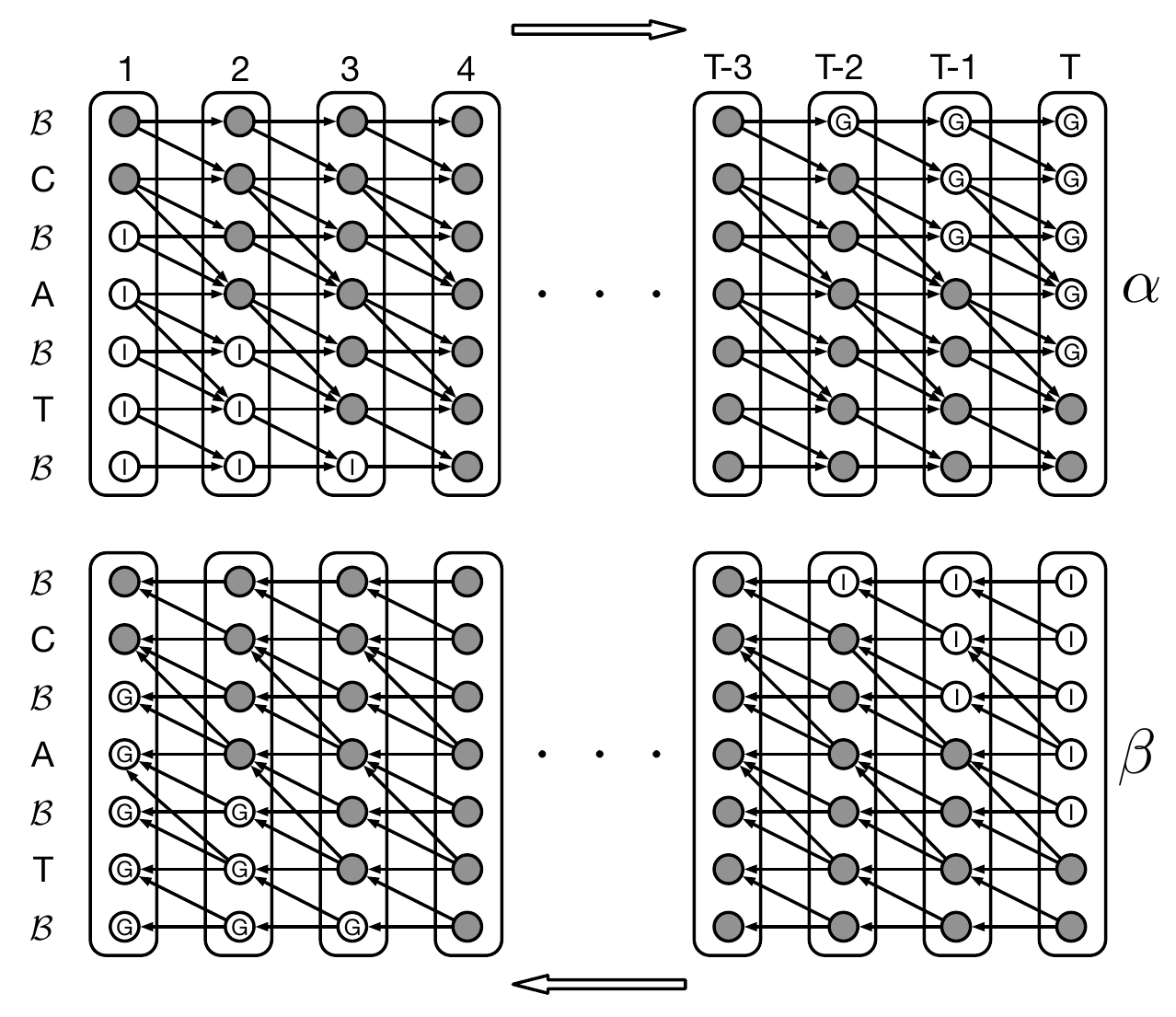}
    \caption{Forward and backward pass for GPU implementation of CTC. Gray circles contain valid values, circle with \textsf{I} contain $-\infty$ and circle with \textsf{G} contain garbage values that are finite. $\mathcal{B}$ stand for the blank character that the CTC algorithm adds to the input utterance label. Column labels on top show different time-steps going from \textsf{1} to \textsf{T}. 
    }
    \label{fig:gpuctc-forward-backward}
\end{figure}

We wrote a GPU-based implementation of CTC in order to overcome these two problems. The key insight behind our implementation is that we can compute all elements in each column of the $\alpha$ matrix, rather than just the valid entries. If we do so, Figure~\ref{fig:gpuctc-forward-backward} shows that invalid elements either contain a finite garbage value (\textsf{G}), or $-\infty$ (\textsf{I}), when we use a special summation function that adds probabilities in $\log$ space that discards inputs that are $-\infty$. This summation is shown in Figure~\ref{fig:gpuctc-forward-backward} where arrows incident on a circle are inputs and the result is stored in the circle. However, when we compute the final gradient by element-wise summing $\alpha$ and $\beta$, all finite garbage values will be added with a corresponding $-\infty$ value from the other matrix, which results in $-\infty$, effectively ignoring the garbage value and computing the correct result. One important observation is that this element-wise sum of $\alpha$ and $\beta$ is a simple sum and does not use our summation function.

To compute the gradient, we take each column of the matrix generated from element-wise addition of $\alpha$ and $\beta$ matrices, and do a key-value reduction using the character as key, using the ModernGPU library~\cite{ModernGPU}. This means elements of the column corresponding to the same character will sum up their values. In the example shown in Figure~\ref{fig:gpuctc-forward-backward}, the blank character, $\mathcal{B}$, is the only repeated character and at some columns, say for $t=1$ of $t=2$, both valid elements (gray) and $-\infty$ correspond to it. Since our summation function in $\log$ space effectively ignores the $-\infty$ elements, only the valid elements are combined in the reduction.

In our GPU implementation, we map each utterance in the minibatch to a CUDA \emph{thread block}. Since there are no dependencies between the elements of a column, all of them can be computed in parallel by the threads in a threadblock. There are dependencies between columns, since the column corresponding to time-step $t+1$ cannot be computed before the column corresponding to time-step $t$. The reverse happens when computing the $\beta$ matrix, when column corresponding to time-step $t$ cannot be computed before the column corresponding to time-step $t+1$. Thus, in both cases, columns are processed sequentially by the thread block. 

Mapping the forward and backward passes to corresponding CUDA kernels is straightforward since there are no data dependencies between elements of a column. The kernel that does the backward pass also computes the gradient. However, since the gradients must be summed up based on the label values, with each character as key, we must deal with data dependencies due to repeated characters in an utterance label. For languages with small character sets like English, this happens with high probability. Even if there are no repeated characters, the CTC algorithm adds $\mathsf{L}+1$ blank characters to the utterance label. 
We solve this problem by performing a key-value sort, where the keys are the characters in the utterance label, and the values are the indices of each character in the utterance. After sorting, all occurrences of a given character are arranged in contiguous segments. We only need to do the sort once for each utterance. The indices generated by the sort are then used to sequentially sum up the gradients for each character. This sum is done once per column and in parallel over all characters in the utterance. Amortizing the cost of key-value sort over $\mathsf{T}$ columns is a key insight that makes the gradient calculation fast.

Our GPU implementation uses fast \emph{shared memory} and registers to achieve high performance when performing this task. Both forward and backward kernels store the $\alpha$ matrix in \emph{shared memory}. Since \emph{shared memory} is a limited resource, it is not possible to store the entire $\beta$ matrix. However, as we go backward in time, we only need to keep one column of the $\beta$ matrix as we compute the gradient, adding element-wise the column of the $\beta$ matrix with the corresponding column of the $\alpha$ matrix. Due to on-chip memory space constraints, we read the output of the softmax function directly from off-chip \emph{global memory}. 

Due to inaccuracies in floating-point arithmetic, especially in transcendental functions, our GPU and CPU implementation are not bit-wise identical. This is not an impediment in practice, since both implementations train models equally well when coupled with the technique of sorting utterances by length mentioned in Section~\ref{subsection:sorting}. 

\end{document}